%% file: main.tex
\documentclass[sigconf, nonacm, 10pt]{acmart}

\usepackage[utf8]{inputenc}
\usepackage{natbib}
\usepackage{graphicx}

\usepackage{caption}
\usepackage{subcaption}
\usepackage{xspace}
\usepackage{makecell}
\usepackage{multicol}
\usepackage{multirow}
\usepackage{tikz}
\usepackage[linesnumbered,ruled,vlined]{algorithm2e}
\usepackage{algpseudocode}
\usepackage{amsmath}
\usepackage{paralist}

\SetCommentSty{mycommfont}

\usepackage{threeparttable}

\usepackage{hyperref}

\newcommand*{\circled}[1]{\lower.7ex\hbox{\tikz\draw (0pt, 0pt) circle (.5em) node {\makebox[1em][c]{\small #1}};}}

\newcommand{\sysname}[0]{{\sc ActiveFlow}\xspace}

\begin{document}

\title{Scaling Up On-Device LLMs via Active-Weight Swapping Between DRAM and Flash}

\author{Fucheng Jia}
\authornote{Research interns at Microsoft Research.}
\affiliation{%
   \institution{Central South University\\Microsoft Research}
   \country{}
  }
\email{fuchengjia@csu.edu.cn}

\author{Zewen Wu\footnotemark[1]}
\affiliation{%
   \institution{Tsinghua University University\\Microsoft Research}
   \country{}
  }
\email{wuzw21@mails.tsinghua.edu.cn}

\author{Shiqi Jiang}
\affiliation{%
   \institution{Microsoft Research}
   \country{}
  }
\email{shijiang@microsoft.com}

\author{Huiqiang Jiang}
\affiliation{%
   \institution{Microsoft Research}
   \country{}
  }
\email{hjiang@microsoft.com}

\author{Qianxi Zhang}
\affiliation{%
   \institution{Microsoft Research}
   \country{}
  }
\email{qianxi.zhang@microsoft.com}

\author{Yuqing Yang}
\affiliation{%
   \institution{Microsoft Research}
   \country{}
  }
\email{yuqing.yang@microsoft.com}

\author{Yunxin Liu}
\affiliation{%
   \institution{Institute for AI Industry Research (AIR), Tsinghua University}
   \country{}
   }
\email{liuyunxin@air.tsinghua.edu.cn}

\author{Ju Ren}
\affiliation{%
   \institution{Tsinghua University}
   \country{}
  }
\email{renju@tsinghua.edu.cn}

\author{Deyu Zhang}
\affiliation{%
   \institution{Central South University}
   \country{}
  }
\email{zdy876@csu.edu.cn}

\author{Ting Cao}
\authornote{Corresponding author.}
\affiliation{%
   \institution{Institute for AI Industry Research (AIR), Tsinghua University}
   \country{}
  }
\email{tingcao@mail.tsinghua.edu.cn}

\pagestyle{empty}








\input{abstract_old}
\maketitle

\input{introduction}

\input{motivation_and_analysis}
\input{design_v2}

\input{implementation}
\input{evaluation}
\input{related_works}
\input{conclusion}

\balance
\bibliographystyle{ACM-Reference-Format}
\bibliography{references}

\end{document}

%% file: abstract_old.tex
\begin{abstract}
Large language models (LLMs) are increasingly being deployed on mobile devices, but the limited DRAM capacity constrains the deployable model size. This paper introduces \sysname, the first LLM inference framework that can achieve adaptive DRAM usage for modern LLMs (not ReLU-based), enabling the scaling up of deployable model sizes. The framework is based on the novel concept of \textit{active weight DRAM-flash swapping} and incorporates three novel techniques: (1) Cross-layer active weights preloading. It uses the activations from the current layer to predict the active weights of several subsequent layers, enabling computation and data loading to overlap, as well as facilitating large I/O transfers. (2) Sparsity-aware self-distillation. It adjusts the active weights to align with the dense-model output distribution, compensating for approximations introduced by contextual sparsity. (3) Active weight DRAM-flash swapping pipeline. It orchestrates the DRAM space allocation among the hot weight cache, preloaded active weights, and computation-involved weights based on available memory. Results show \sysname achieves the performance-cost Pareto frontier compared to existing efficiency optimization methods.    
\end{abstract}

%% file: introduction.tex
\section{Introduction}
Large language models (LLMs) are increasingly deployed on mobile and PC devices as integral system components, such as the on-device 3B Apple foundation model for Apple iOS~\cite{appleincIntroducingApplesOnDevice}, the 3.82B Phi Silica for Windows~\cite{phi-silica}, and 3.35B Gemini Nano for Google's Android~\cite{teamGeminiFamilyHighly2024}. 

However, further scaling up the on-device LLM size is very difficult, with a key constraint of DRAM size. Due to power and area constraints, the DRAM size on mobile devices remains limited and difficult to increase, even across device upgrades (e.g., both iPhone 15 and iPhone 16 feature 8\,GB DRAM). Furthermore, the available DRAM capacity is also determined by the co-active apps and OS processes remaining in DRAM simultaneously. Mobile OS can terminate an app under low available DRAM unless the app can reduce the memory usage\cite{reduce_mem}. 

\textbf{Goal.} To enable the deployment of larger LLMs, it is essential to realize \textit{adaptive DRAM usage} for LLM inference. That is, the inference process dynamically adapts to different available DRAM sizes while maintaining comparable model quality and inference speed. 
Mirroring the OS employs virtual memory to abstract physical limitations, this work aims for adaptive DRAM usage that is transparent to the user, creating the illusion that the entire model resides in DRAM.

Adaptive DRAM usage has been previously investigated for traditional non-autoregressive DNNs (e.g., CNN and Bert) through DRAM-Flash swapping~\cite{flexnn,coldstart}. However, the fundamental difference in workload characteristics hinders the direct application of these methods to LLMs. Existing techniques rely on the computation-intensive feature of traditional DNNs, so the current operator computation can overlap the loading of the next operator. While this overlap is present in the LLM prefilling stage, the significantly more time-consuming autoregressive decoding phase is bottlenecked by memory access. Consequently, realizing user-oblivious adaptive memory management for LLM inference necessitates minimizing Flash data loading to mitigate the substantial disparity between memory and Flash bandwidth ($\sim5\times$ on mobile phones). 



\begin{figure}[t]
    \centering
    \begin{subfigure}[t]{0.48\textwidth}
        \centering
        \includegraphics[width=0.95\linewidth]{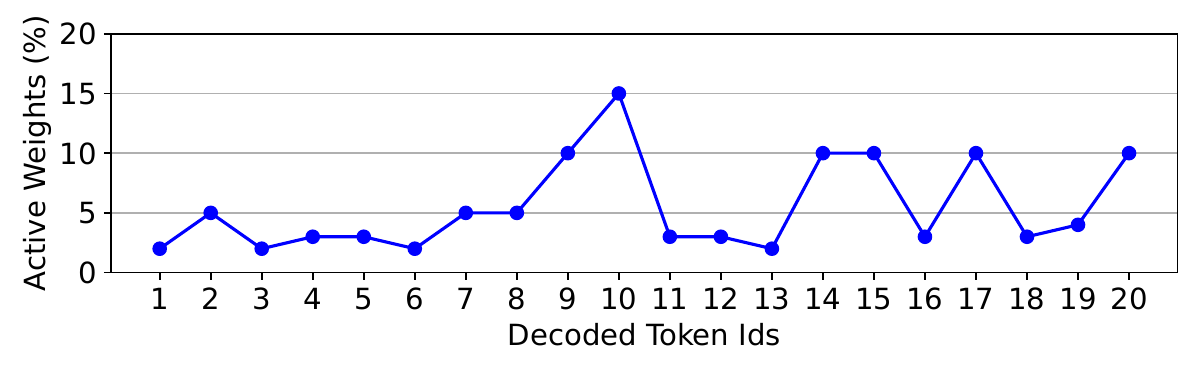}
    \end{subfigure}
    \vspace{-1em}
    \caption{The upper bound sparsity of LLaMa-2-70B model during decoding.}
    \label{F.upper_bound_sparsity}
\end{figure}

Fortunately, a unique characteristic of LLMs is \textit{contextual sparsity}, where although the model itself is large, only a small subset of weights is actively used per token generation~\cite{pmlr-v202-liu23am}, which we term as \textit{active weights}. Our upper-bound analysis (Fig.~\ref{F.upper_bound_sparsity}) shows that during each inference iteration, only <15\% weights need to be activated to generate the same token. 

\begin{figure}[t]
    \centering
    %
    \includegraphics[width=1\linewidth]{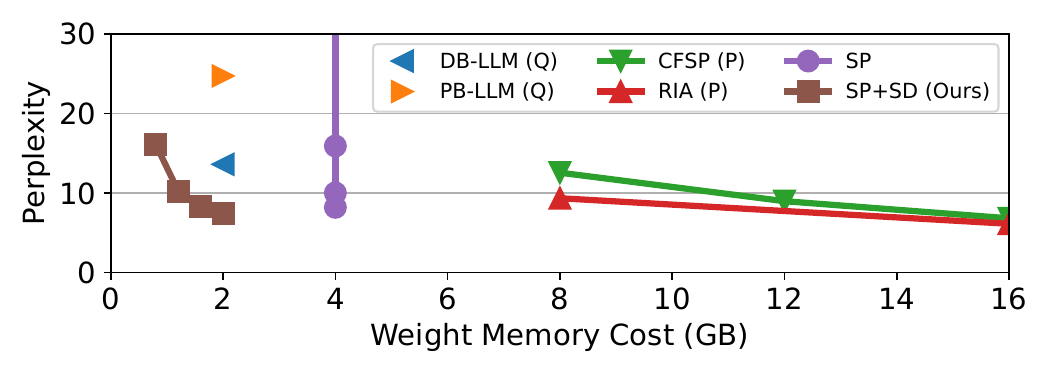}
    \vspace{-1em}
    \caption{The perplexity versus cost of LLaMA-3-8B model. Ours shows the Pareto frontier compared with SOTA model compression methods including quantization (Q), pruning (P) and contextual sparsity (SP). Each point on the scaling line means a sparsity ratio.}
    \label{F.pareto_front}
\end{figure}

\textbf{Challenges.} This contextual sparsity inspires us explore the new opportunity of \textit{active weights swapping} for adaptive memory usage. Unlike traditional per-operator swapping, active weight swapping introduces greater challenges: (1) How to accurately identify the active weights, given contextual sparsity is highly dynamic, varying cross tokens, layers and blocks. Misidentification could degrade model accuracy. (2) How to predict the active weights as early as possible, allowing for overlapping computation with loading, as well as efficient large I/O transfers, both of which are critical for performance.    

\begin{figure}[!t]
    \centering
    \begin{subfigure}[t]{0.48\textwidth}
        \centering
        \includegraphics[width=0.95\linewidth]{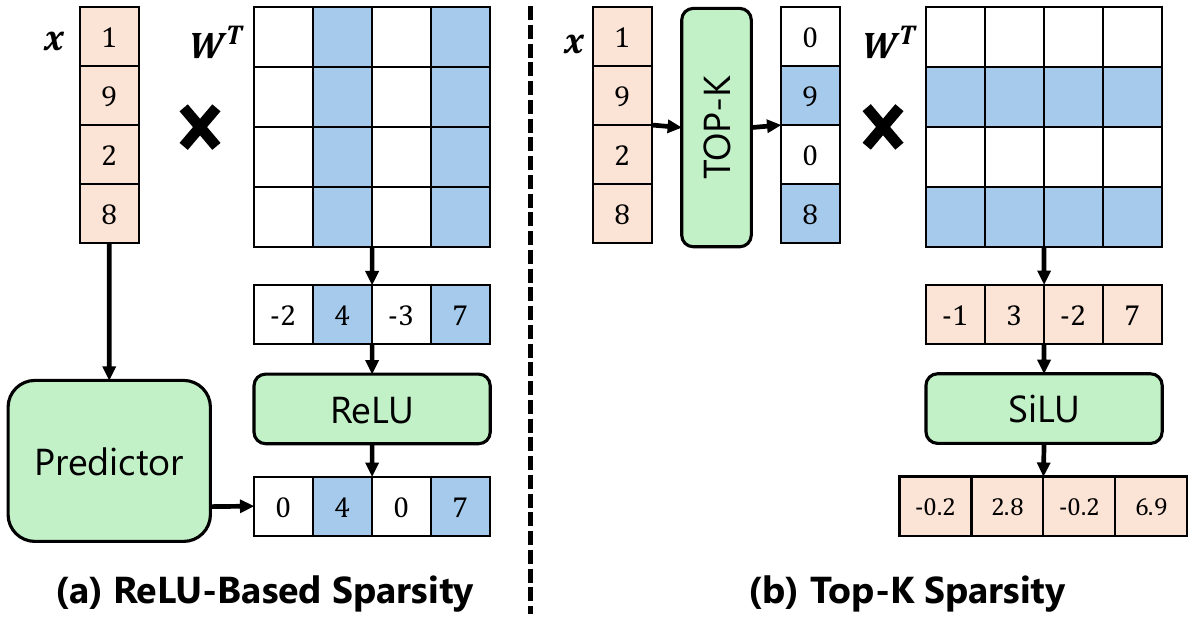}
    \end{subfigure}
    \vspace{-1em}
    \caption{
    The ReLU-based sparsity and Top-K activation sparsity.   
    We base our system on Top-K sparsity due to its broader applicability and higher accuracy.}
    \label{F.dynamic_sparisties}
\end{figure}

Several works have explored contextual sparsity\cite{pmlr-v202-liu23am,powerinfer-2,10.1145/3694715.3695964,2024q-sparse, l2024hirehighrecallapproximate,mirzadeh2023relustrikesbackexploiting,turbosparse,prosparse}, but gaps remain in addressing the challenges above. Some methods like Deja Vu~\cite{pmlr-v202-liu23am}, PowerInfer~\cite{10.1145/3694715.3695964} and LLM in a flash~\cite{llm_in_a_flash} use available ReLU-based models to generate zero activations and introduce additional predictors (GB memory cost) to forecast these zeros.  However, modern LLMs used in productions (e.g., LLaMA) rarely use ReLU-based architecture due to its inferior  accuracy~\cite{touvron2023llama} (see Fig.\ref{F.end_to_end_ppl_and_memory_cost}).
There are also works performing continued pre-training to transform available models to ReLU or ReLU-variant based, such as PowerInfer-2~\cite{powerinfer-2}, TurboSparse~\cite{turbosparse}, ProSparse~\cite{prosparse}, and Q-Sparse~\cite{2024q-sparse}. These works require training on hundreds of billions of tokens and consume substantial hardware resources. There are also works, such as InfiniGen~\cite{infinigen}, NSA~\cite{nsa}, and SeerAttention~\cite{seerattention}, focusing on KV cache sparsity but not weight sparsity. These methods benefit long context scenarios (>32K) which are not the common cases on edge. TEAL~\cite{liu2025trainingfreeactivationsparsitylarge} proposes a training-free, magnitude-based sparsity method (see Fig.~\ref{F.dynamic_sparisties}), where only activations above a threshold are computed. However, the active weights cannot be predicted, but only be identified after the input activation is ready. Additionally, the method is empirical, and there is no mechanism to compensate for the accuracy loss due to the potential  misidentification of active weights.
Therefore, current techniques fall short of achieving adaptive memory usage for LLMs.


\textbf{Our work.} This paper proposes \sysname LLM inference framework.  It can realize user-oblivious adaptive DRAM usage, in order to scale up the LLM sizes that can be deployed on mobile devices. Similar to TEAL, this paper utilizes magnitude-based, model-architecture-independent activation sparsity, to ensure the framework's applicability to modern LLMs. 
Beyond that, \sysname incorporates three novel techniques. 

Firstly, \textbf{Cross-layer active weight preloading}. To address the sequential dependency issue of active weights with its input activation in order to enable computation and loading overlapping, we propose cross-layer active weight preloading. It creatively utilizes the current layer's activation to pre-identify the next n layers' active weights. It is based on the obeservation that due to the widely used residual connection, the activation magnitude distribution across layers share significant similarity (>80\% shown in Fig.~\ref{F.cosine_similarity_and_top_k_precision}). For the active weights that missed by pre-loading, \sysname loads on-demand when the actual activation is ready.

Secondly,  \textbf{Sparsity-aware self-distillation.} 
Even the magnitude-based activation sparsity empirically has shown the superior quality compared to other sparse methods~\cite{liu2025teal}, it still introduces an approximation compared to the dense model. To compensate for the approximation, we propose sparse-aware self-distillation to adjust the active weights towards the dense-model output. The distillation improves both the sparsity ratio and model accuracy. The technique is inspired by and integrated with the quantization-aware self distillation~\cite{du2024bitdistiller}. Similar to this work, the self-distillation only needs several A100 GPU hours to train. The two methods can be used collaboratively for LLM deployment. 

Thirdly, \textbf{DRAM-flash active weight swapping pipeline.} The pipeline reorganizes the data layout for the cross-layer preloading, and overlaps the active weight loading with the current layer computing. It also integrates a contextual hot active weight caching policy beyond naive swapping. The pipeline orchestrates the space allocations among the cache, preloaded active weights, and computation involved weights  according to available memory. 

We implement \sysname and evaluate it on different mobile phones (OnePlus 12, Pixel 6, and Infinix Zero). Results (Fig.~\ref{F.pareto_front}, more in Sec.~\ref{evaluation}) show that \sysname achieves \textbf{the inference performance-cost Pareto frontier} among existing efficiency optimization methods, including state-of-the-art quantization (DB-LLM~\cite{chen2024dbllmaccuratedualbinarizationefficient} and PB-LLM~\cite{shang2023pbllmpartiallybinarizedlarge}), pruning (CPSP~\cite{wang2024cfspefficientstructuredpruning} and RIA~\cite{yi2025symmetricpruninglargelanguage}), and contextual sparsity (TEAL~\cite{liu2025trainingfreeactivationsparsitylarge}), demonstrating its practical value. Particularly, under the same model quality and speed, \sysname reduces the DRAM usage by up to 40\% for LLaMA 7B compared to llama.cpp. Under the same sparsity ratio, \sysname can reduce memory by 2$\times$ compared to TEAL. \sysname is the first to successfully deploy the original Mixtural-8x7B 4bit model~\cite{jiang2024mixtralexperts} (no ReLU introduced) on a mid-range pixel-6 phone, achieveing 1.8 tokens/s with 2.9\,GB memory cost.  


To summarize, the contributions of this paper are: 
\begin{itemize}
  \item We propose \sysname, the first LLM inference system to enable user-oblivious adaptive DRAM usage through active weight swapping for modern general LLMs without ReLU dependency. 
  \item We propose the cross-layer active weights preloading to allow computation/loading overlapping and large I/O transfer. 
  \item We propose sparsity-aware self distillation to compensate the approximation introduced by sparsity. 
  \item We implement the end-to-end \sysname. Results show it achieves the inference quality-cost Pareto frontier among existing optimization methods. 
\end{itemize}

%% file: motivation_and_analysis.tex
\section{Motivation and Background}


\subsection{Upper Bound Analysis of Contextual Sparsity in LLMs}

A specific feature of LLMs is \textit{contextual sparsity}~\cite{pmlr-v202-liu23am,powerinfer-2,10.1145/3694715.3695964,2024q-sparse, l2024hirehighrecallapproximate,mirzadeh2023relustrikesbackexploiting,turbosparse}, which means a small, context-dependent subset of total weights, that can generate the same output as the full model. We term this small subset of weights as \textit{active weight}. Compared to the static sparsity from model pruning~\cite{sparsegpt,wander}, contextual sparsity dynamically selects different active weights for computation during each token generation, preserving the model’s overall capacity and adaptability. Contextual sparsity has also been empirically demonstrated to be compatible with model quantization~\cite{2024q-sparse}. 

Since our techniques will be based on contextual sparsity, we first analyze the upper bound of this sparsity.  
We use a Llama-2-70B model to evaluate the amount of active weights required to generate the same token with full weights during the decoding process. The evaluation is conducted by incrementally removing unimportant weights for each decoded token by 1\%. The important scores of weights are calculated by 
$S_{ij}=|W_{ij}| \cdot |X_{j}|$ , where $W_{ij}$ is an element of weight matrix and $X_{j}$ is an element of the input activation vector. As shown in Fig. \ref{F.upper_bound_sparsity}, the results indicate that most tokens require less than 5\% of the weights, with the maximum active weight being only 15\%. This high level of sparsity shows a great potential for reduced inference cost.  

\begin{figure}[!t]
    \centering
    \begin{subfigure}[t]{0.43\textwidth}
        \centering
        \includegraphics[width=\textwidth]{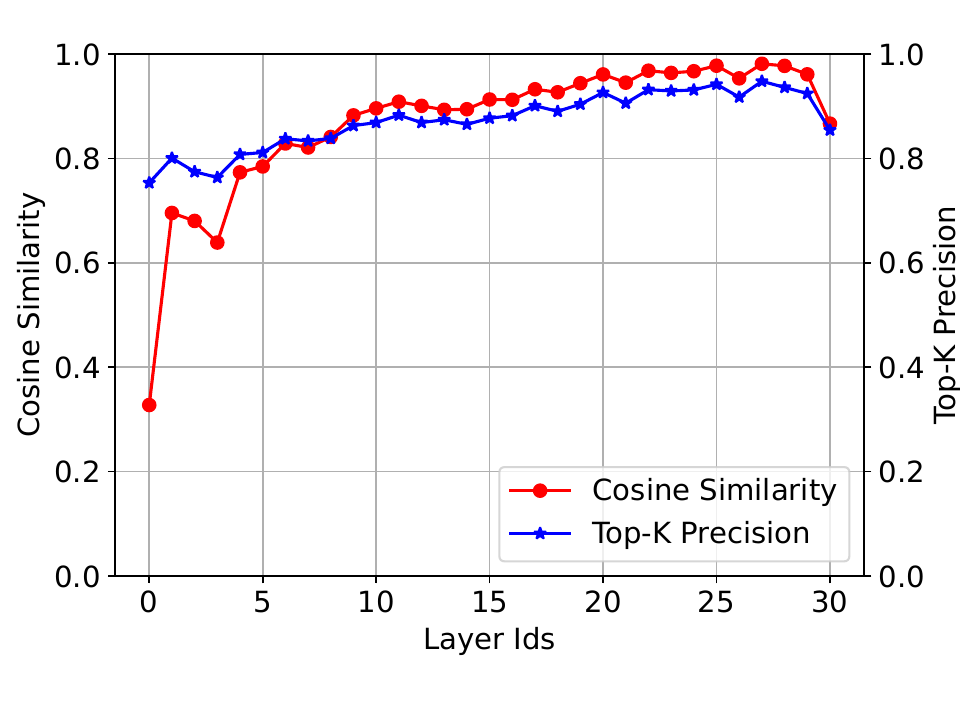}
        \caption{}
        \label{F.cosine_similarity_and_top_k_precision}
    \end{subfigure} \\
    \begin{subfigure}[t]{0.47\textwidth}
        \centering
        \includegraphics[width=\textwidth]{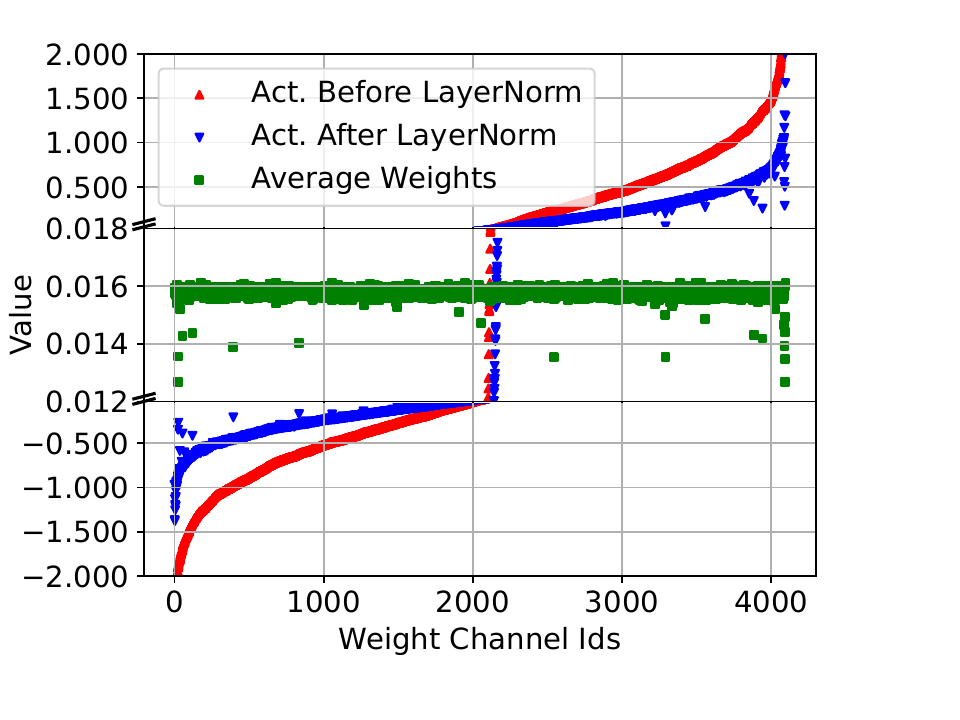}
        \caption{}
        \label{F.activation_before_after_layernorm}
    \end{subfigure}
    \caption{The cross-layer input activation similarity of a LLaMA-2-7B model. (a) The attention input cosine similarity and Top-K precision. (b) The value of activation before/after LayerNorm layer and average weights.}
    \label{F.cross_layer_activation_similarity}
\end{figure}


Although the above results are promising, it is challenging to identify the active weights during inference, unless the weights are loaded and computed with activations. 
Consequently, some works~\cite{pmlr-v202-liu23am,prosparse} rely on ReLU-generated sparsity and propose extra predictors to estimate the sparsity, as illustrated in Fig.~\ref{F.dynamic_sparisties}(a). These predictors are trained with calibration datasets, loaded into memory, and executed before performing per-layer LLM computations. However, the deployment cost of predictors is significant because (1) the datasets may not be suitable for real user data, (2) predictors require additional memory (at the GB level), and (3) they introduce extra computational overhead.

More recent works~\cite{liu2025teal,2024q-sparse} propose magnitude-based activation sparsity, as shown in Fig.~\ref{F.dynamic_sparisties}(b). We term this sparsity as \textit{Top-K sparsity} following~\cite{2024q-sparse}. Only the activation elements with a magnitude above a threshold will be computed for each operator. Top-K sparsity demonstrates obvious advantages: 1) compatibility to modern non-ReLU LLMs; 2) applicability to all linear transformation operators rather than just FFN blocks; 3) no extra predictors needed. 

These advantages motivate us to identify active weights for swapping based on Top-K activation sparsity.

\subsection{Observation: Similarities in Cross-Layer Activations}

\begin{figure}[tb]
    \centering
    \begin{subfigure}[t]{0.48\textwidth}
        \centering
        \includegraphics[width=0.95\linewidth]{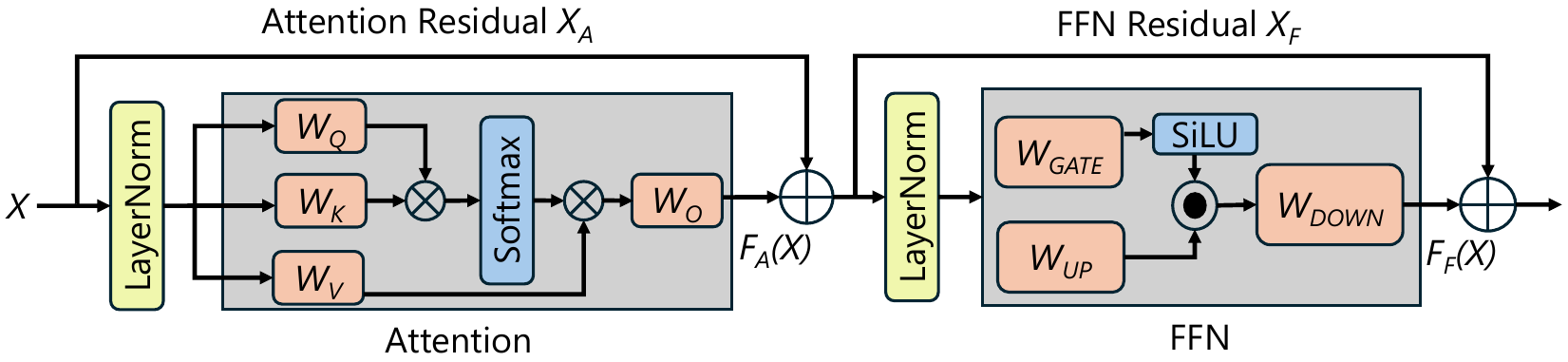}
    \end{subfigure}
    \vspace{-1em}
    \caption{The simplified transformer layer structure of an LLM model. Residual connections pass the input of a block directly to output.} 
    \label{F.decoder_module}
\end{figure}

A key observation of this paper is that \textbf{the input activations of the attention and MLP blocks in LLMs exhibit high cross-layer similarity}. Fig. \ref{F.cosine_similarity_and_top_k_precision} uses the input activation of the attention block as an example to show the cosine similarity and Top-K sparsity precision in each consecutive two layers in a Llama-2-7B model. Starting from the 3rd layer, the attention Q, K, V, and FFN gate and up operators exhibit over 95\% similarity. Consequently, the Top-K sparsity precision for these operators exceeds 80\% cross layers.

The similarity is primarily due to the significant contribution of the residuals to the input activations. Fig. ~\ref{F.decoder_module} shows a simplified transformer layer structure. The input activations are composed of the sum of two elements: the output activation of the previous block $F(X)$ and the residual $X$. 
The cross-layer similarity is because the residual values $X$ are larger than the output activation values $F(X)$. This difference in values arises from (1) the LayerNorm layer in the attention and MLP blocks, and (2) the weights magnitudes. As shown in Fig. \ref{F.activation_before_after_layernorm}, the LayerNorm reduces the activation magnitude by 50\%. Additionally, the weight magnitude is smaller than the activation magnitude, resulting in a smaller calculation output.

The cross-layer input similarity motivates us for cross-layer preloading, which uses current layer's activation to identify following layers' active weights.  


\subsection{Observation: Contextual Hot Active Weights During Decoding}

\begin{figure}[!t]
    \centering
    \includegraphics[width=0.55\textwidth]{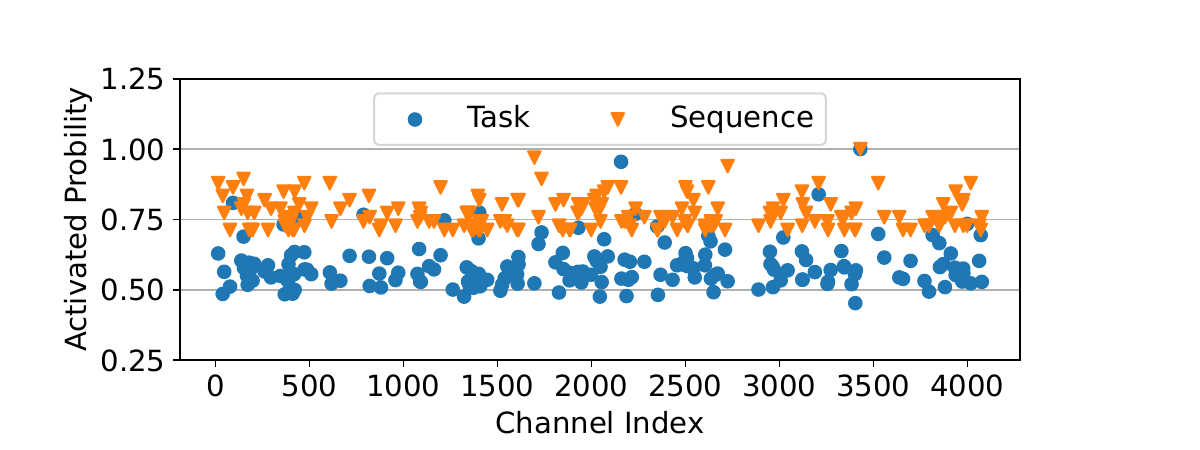}
    \caption{The selection probability of active weights in attention Q/K/V operators of Llama-2-7B model (under 50\% contextual sparsity). Context level shows higher selection probability than task level.
    We only show the active weight with probability > 0.7.}
    \label{F.activated_prob_of_task_and_sequence}
\end{figure}

This section investigates the presence of \textit{hot active weights}, i.e., the weights that are frequently selected across inference iterations during decoding. This investigation aims to identify opportunities for caching and more intelligent swapping strategies. Our observation is that \textbf{contextual active weights exhibit high temporal locality across inference iterations during decoding}, suggesting that caching hot active weights for higher cache hit rates. 

As shown in Fig.~\ref{F.activated_prob_of_task_and_sequence}, we conducted two levels of active weight selection frequency analysis: \textit{task level} and \textit{context level}. The task level counts the frequency with which weight channels are selected during the decoding process for all input contexts across a dataset (WikiText-2). In contrast, the context level counts the frequency of weight selection specifically for the decoding process of a given input context. Results show that hot weight selection probabilities on the context level exceeds 0.7, while the task level exceeds 0.5. The difference demonstrates the potentially improved cache hit and reduced loading cost by implementing a contextual cache management policy.  

%% file: design_v2.tex
\section{Cross-layer Active Weight Preloading}

\begin{figure}[!t]
    \centering
    \begin{subfigure}[t]{0.48\textwidth}
        \centering
        \includegraphics[width=0.95\linewidth]{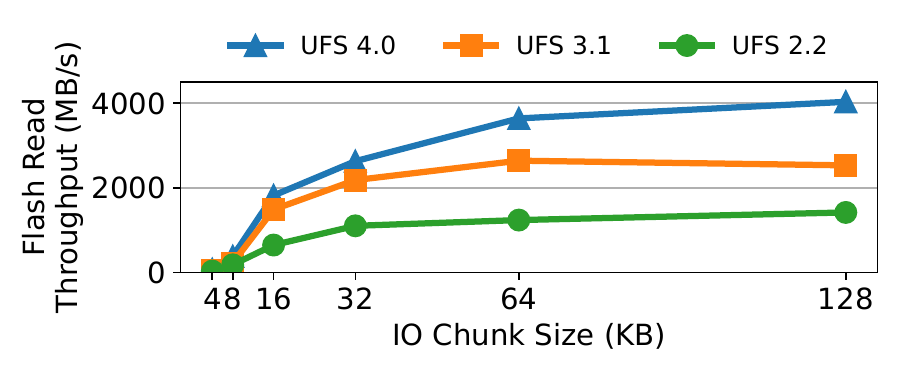}
    \end{subfigure}
    \vspace{-1em}
    \caption{The flash read throughput of various IO chunk sizes on three devices with difference UFS capabilities.}
    \label{F.chunk_size_and_flash_bandwidth}
\end{figure}

To realize adaptive DRAM usage, two critical challenges for performance is: (1) whether the weight loading and computation can be overlapped to hide the flash loading overhead; (2) whether the I/O transfer can fully utilize the flash bandwidth. As shown in Fig.~\ref{F.chunk_size_and_flash_bandwidth}, the flash read throughput varies greatly with the chunk size of each I/O transfer. To achieve the peak flash throughput, the chunk size has to >64\,KB. However, active weight from Top-K activation sparsity is in channel granularity, e.g., 4KB (see Fig.~\ref{F.dynamic_sparisties}), and naive loading of the each active weight channel from flash can reduce the throughput from GB/s to MB/s.     

However, current works including PowerInfer~\cite{powerinfer-2,10.1145/3694715.3695964}, LLM in Flash~\cite{llm_in_a_flash} and Ripple~\cite{wang2024rippleacceleratingllminference} only partially alleviated the problem. To enlarge the chunk size, they cluster co-active weight channels within the same block, and overlap each cluster loading and computation.  

\begin{figure}[t]
    \centering
    \begin{subfigure}[t]{0.48\textwidth}
        \centering
        \includegraphics[width=0.95\linewidth]{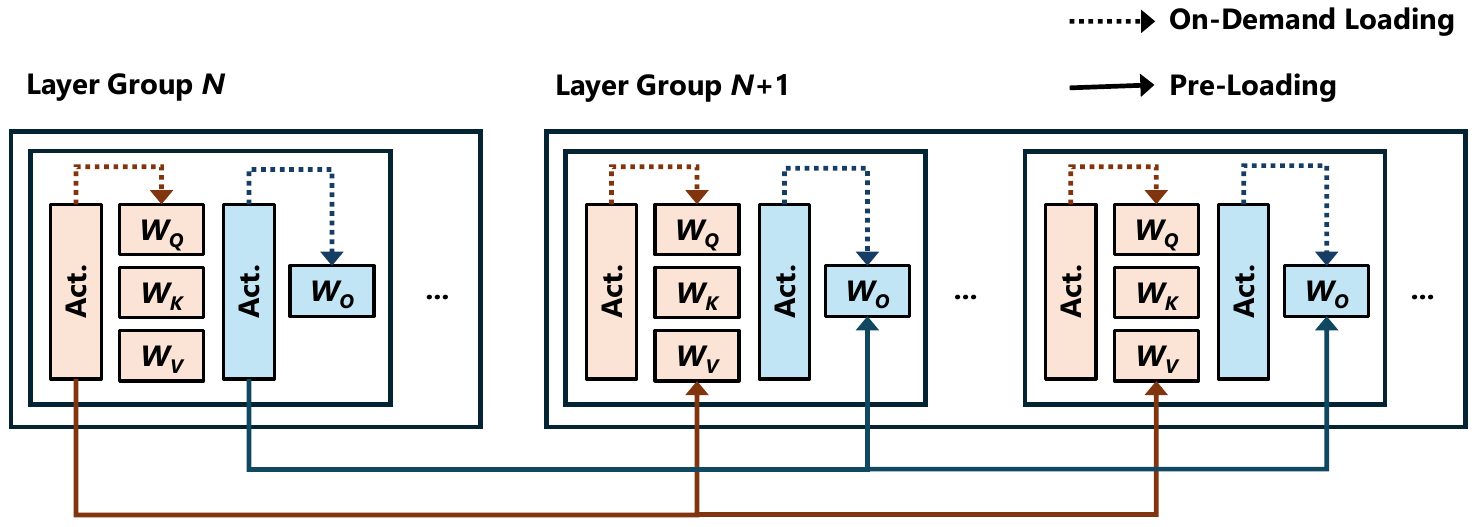}
    \end{subfigure}
    \vspace{-1em}
    \caption{Cross-layer active weight pre-loading. While the computing of current layer, the active weights of all the operators in the next N layers (layer group) will be preloaded based on the current activation. The missed active weights during preloading will be on-demand loaded after its actual activation is ready.}
    \label{F.cross_layer_pipeline}
\end{figure}

\textbf{Our technique.} To overcome the challenges, based on our key observation that cross-layer activations exhibit significant similarity, we propose the cross-layer active weight preloading. As shown in Fig.~\ref{F.cross_layer_pipeline}, while the computing of current layer, the next N layers' active weights will be preloaded to DRAM simultaneously. We term these N layers as a \textit{layer group} for preloading.  The N is set based on the available DRAM, and the computing latency (N=4 can fully overlap the loading and computing in our evaluation). The preloading will include the active weights from all the operators in both Attention and FFN blocks. Different activations correspond to different parts of the weights being loaded. For example, Q, K, and V activations are only used to load $W_q$, $W_k$, and $W_v$, respectively. 

\begin{figure}[t]
    \centering
    \begin{subfigure}[t]{0.48\textwidth}
        \centering
        \includegraphics[width=\linewidth]{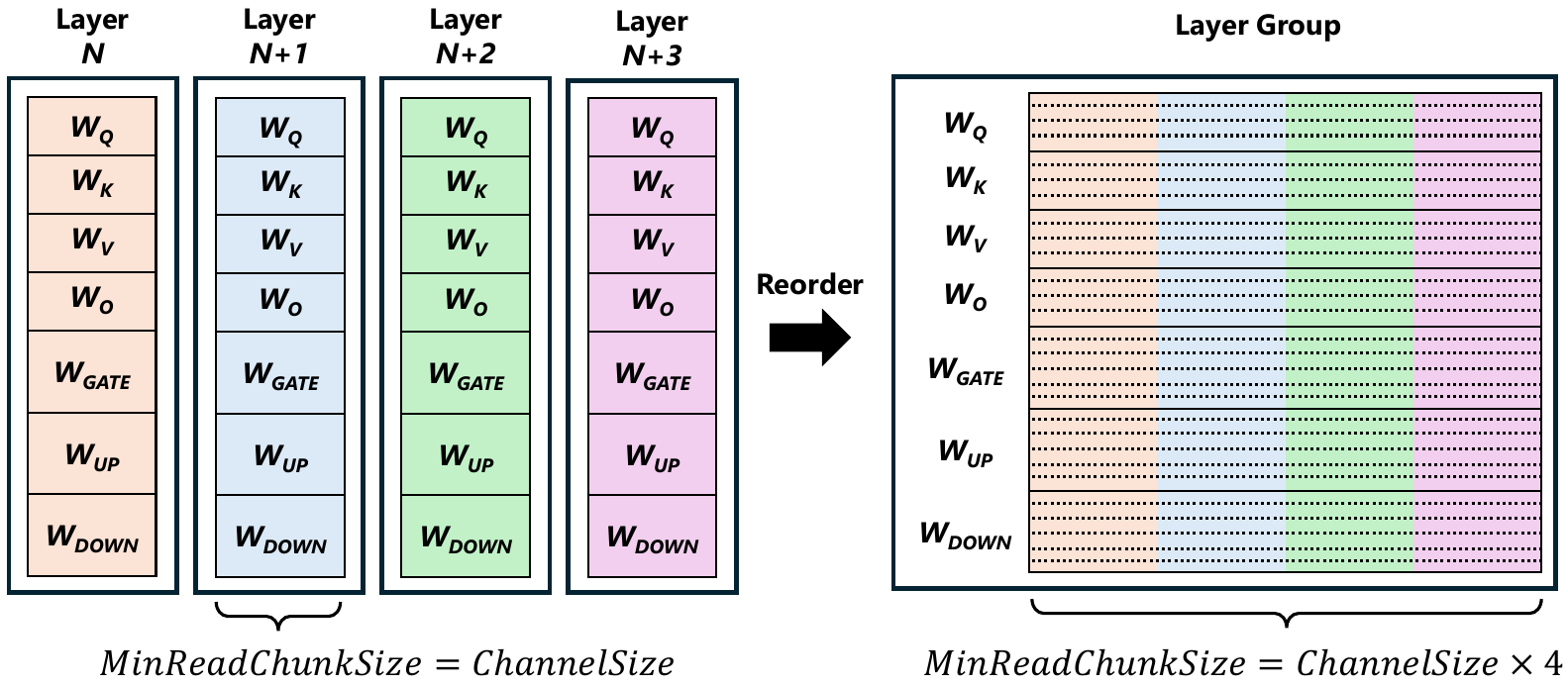}
    \end{subfigure}
    \vspace{-1em}
    \caption{The reordered weights in a 4-layers group. The weight layout now is in the order of weight channel, layer, and operator type. By multi-layer weight reordering, the minimal loading chunk size is increased to improve the loading efficiency.}
    \label{F.weight_order_in_layer_group}
\end{figure}

Since cross-layer activation similarity is not 100\%, pre-loading can only load a portion of the necessary weights in advance. Any remaining weights that were not correctly pre-loaded are fetched through on-demand loading.  This only takes $\sim5\%$ of the total active weights.  

\textbf{Data layout. } To facilitate the cross-layer preloading, the weight layout in flash is reordered, to break the tensor and layer boundary.  As shown in Fig. \ref{F.weight_order_in_layer_group} (left), the normal LLM weight layout is to arrange each weight tensor sequentially for all the operators within each layer. It is inefficient for channel-wise active weight loading. Our approach reorders the weight channels within a preloading layer group according to the order of the channel ID, layer ID, and operator type. For example, $W_q$ weight layout in the layer group is [$Ch\,0_{layerN}$, $Ch\,0_{layerN+1}$, $Ch\,0_{layerN+2}$, $Ch\,0_{layerN+3}$, $Ch\,1_{layerN}$, $Ch\,1_{layerN+1}$, $Ch\,1_{layerN+2}$, $Ch\,1_{layerN+3}$, ..]. 
This reordering enables pre-loading multiple layers’ weights for the same channel in a single read operation, significantly increasing the loading chunk size and improving loading efficiency.

\begin{figure}[t]
    \centering
    \begin{subfigure}[t]{0.48\textwidth}
        \centering
        \includegraphics[width=0.95\linewidth]{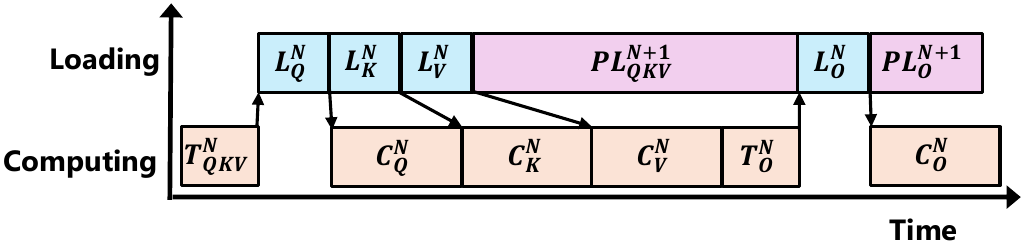}
    \end{subfigure}
    \vspace{-1em}
    \caption{The computing-loading overlap pipeline for LLM inference in an attention block after warming up.}
    \label{F.overlap_pipeline}
\end{figure}


\begin{figure}[t]
    \centering
    \begin{subfigure}[t]{0.48\textwidth}
        \centering
        \includegraphics[width=0.6\linewidth]{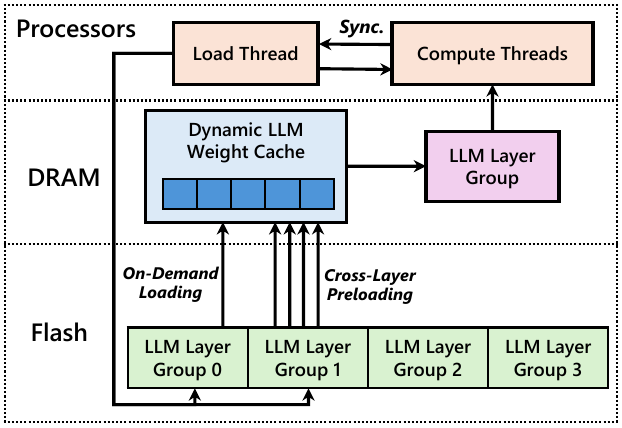}
    \end{subfigure}
    \vspace{-1em}
    \caption{The weight layout and flow of \sysname.}
    \label{F.fig_system_overview}
\end{figure}

\section{Active Weight Swapping Pipeline}

Building on the proposed cross-layer-group LLM weight loading and reordering techniques, we design a LLM computing-loading overlapping execution pipeline as shown in Fig. \ref{F.overlap_pipeline}. The pipeline consists of four main operations: (1) Computing (C) – Performs the required computations. (2) Top-K (T) – Extracts the Top-K mask from activations to determine the indices of the activated weight channels. (3) On-demand loading (L) – Loads weights for the current layer group. (4) Preloading (PL) – Preloads weights for the next layer group.

Fig.~\ref{F.fig_system_overview} demonstrates the weight layout and flow with the pipeline. The whole model resides in the flash with the cross-layer group layout. The current active weights, as well as the pre-loaded and cached weights store in the DRAM. The computation and loading are concurrently executed.    

The overlapped LLM execution pipeline follows two key principles: (1) Maximize the overlap between loading and computing to minimize idle time (bubbles) to fully utilize the memory bandwidth and computing power simultaneously. (2) Maximize the cache hit rate on the sequence level. The challenge is how to accurately estimate the impact of system parameters, such as sparsity, memory cost and cache size on the accuracy and latency of model inference. 



\begin{table}[!t]
  \small
  \caption{The symbols of our system cost model.}
  \label{T.symbol_of_cost_model}
  \begin{tabular}{ll}
    \toprule
    Symbols & Description \\
    \midrule
    \textbf{$sp$} & sparsity of LLM \\
    $hr$ & average hit rate of weight cache \\
    $si$ & average similarity of cross-layer group \\
    \midrule
    $BW_{mem}$ & bandwidth of memory \\
    $BW^{small}_{flash}$ & bandwidth of small chunk reading from flash \\
    $BW^{large}_{flash}$ & bandwidth of large chunk reading from flash \\
    \midrule
    $S_{m}$ & Size of LLM \\
    $S_{l}$ & Size of a LLM layer \\
    \textbf{$N$} & Layer number of a cross-layer group \\
    \midrule
    $M$ & Memory cost of pipeline \\
    $M_{max}$ & Memory budget \\
    $M_{cl}$ & Memory of a cross-layer group \\
    \textbf{$M_{cache}$} & Memory of weight cache \\
    $M_{kv}$ & Memory of KV cache \\
    \midrule
    $T_{decode}$ & Decoding time of a token \\
    $T_{load}$ & Loading time of a cross-layer group \\
    $T_{comp}$ & Computing time of a cross-layer group \\
    $T_{overlap}$ & Overleaping time of two cross-layer groups \\
    $T_{onload}$ & On-demand loading time of a cross-layer groups \\
    $T_{preload}$ & Preloading time of a cross-layer groups \\
  \bottomrule
\end{tabular}
\end{table}

\subsection{Elastic and Optimized LLM Execution}
The goal of this technique is to determine the optimal system parameters, including LLM sparsity, layer number of a cross-layer group, and cache size, for a given mobile device (i.e., with specific computational power and memory budget) and a given LLM. The objective is to minimize system latency while  respecting the memory constraint.

There is tradeoff between LLM sparsity, layer number of a cross-layer group and cache size on the inference metrics in terms of both latency and accuracy. Optimizing one metric could worsen another. To capture this, we define the following problem, with the memory cost as a hard constraint and the objective to minimize the decode latency. The related symbols are listed in Table \ref{T.symbol_of_cost_model}.
\begin{align}
Minimize\ \ \ \ T_{decode} &= T_{load} + T_{overlap} + T_{comp} \label{E.1} \\
M &\le M_{max} \label{E.2}
\end{align}

The decode latency consists of three components: the first cross-layer-group loading time $T_{load}$, the cross-layer-group overlapping time $T_{overlap}$, and the final cross-layer-group computing time $T_{load}$, as in Eq. \ref{E.1}. The loading time $T_{load}$ is the weights missed in the cache divided by the flash loading bandwidth as $BW^{small}_{flash}$, as in Eq. \ref{E.3}. The final cross-layer-group computing time $T_{comp}$ is the group memory size $M_{cl}$ divided by the memory bandwidth $BW_{mem}$, as in Eq. \ref{E.4}. Furthermore, the overlapping time consists of two parts, i.e, the on-demand loading time $T_{load}$ and preloading latency $max(T_{preload}, T_{comp})$, as in Eq. \ref{E.5}. We load the weights that are dissimilar across layers but not present in the cache, with latency $T_{load}$, as in Eq. \ref{E.6}. These weights typically have small chunk sizes, leading to lower bandwidth $BW^{small}_{flash}$. Preloading, on the other hand, loads weights at the cross-layer-group level, fetching only the cache-miss weights (Eq. \ref{E.7}). Since the chunk size in this stage is relatively large, the reading efficiency is significantly higher with bandwidth $BW^{large}_{flash}$.

\begin{align}
    T_{load} &= \frac{M_{cl} \cdot (1-hr)}{BW^{small}_{flash}} \label{E.3} \\
    T_{comp} &= \frac{M_{cl}}{BW_{mem}} \label{E.4} \\
    T_{overlap} &= T_{onload} + max(T_{preload}, T_{comp}) \label{E.5} \\
    T_{onload} &= \frac{S_l \cdot (1-sp) \cdot (1-hr) \cdot (1-si)}{BW^{small}_{flash}} \label{E.6} \\
    T_{preload} &= \frac{M_{cl} \cdot (1-hr)}{BW^{large}_{flash}} \label{E.7}
\end{align}

The memory cost also consists of three components: cross-layer group memory $M_{cl}$, weight cache memory $M_{cache}$, and KV cache memory $M_{kv}$ (Eq. \ref{E.8}). For the KV cache, we only consider the fixed-size case. Therefore, only the first two components will dynamically influence the memory cost. The cross-layer group memory is the size of active weights, as in Eq. \ref{E.9}.


\begin{align}
    M &= M_{cl} + M_{cache} + M_{kv} \label{E.8} \\
    M_{cl} &= S_{l} \cdot (1-sp) \cdot N \label{E.9}
\end{align}

\textbf{Preload-and-computation-balanced cross-layer group search.} We determine the parameters ($sp$, $S_{cl}$, and $M_{cache}$) in a greedy manner, as follows. First, since LLM accuracy is only related to LLM sparsity, we set LLM sparsity by $sp = 1 - (M_{max}/S_{m})$ to ensure the highest accuracy. Second, minimize the decode time recursively. We increase layer number of cross-layer group $L$ in a step by step manner. This brings lower $T_{preload}$. In case $T_{preload} \le T_{comp}$, then stop. Furthermore, if the $T_{preload}$ decrement is less than a threshold, then stop.

This approach ensures near-full memory utilization, minimal latency, and high accuracy. In case that the memory budget changes in online phase, we tune cache size to maintain well overlap between computation and flash read operations.

\subsection{Dynamic LLM Weight Caching}

\begin{figure}[tb]
    \centering
    \begin{subfigure}[t]{0.48\textwidth}
        \centering
        \includegraphics[width=1.00\linewidth]{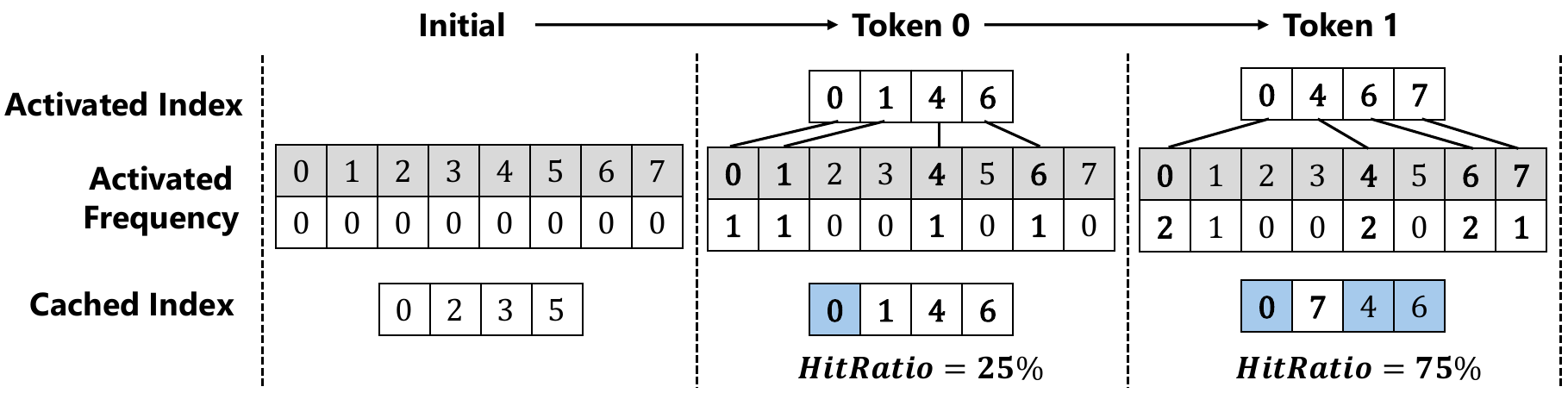}
    \end{subfigure}
    \vspace{-1em}
    \caption{An example of dynamic weights caching during LLM decoding. There are 8 channels in a weight but only half of channels are cached in memory.}
    \label{F.dynamic_cahce}
\end{figure}

To further reduce the number of loaded weights, we design the dynamic LLM weight caching based on observations of hot weights, as illustrated in Fig. \ref{F.dynamic_cahce}. To maximize the cache hit rate, we track the frequency statistics of activation and evict the least-used weights in online phase.

To manage weight eviction, we maintain independent counters for the weights of each layer, ensuring a balanced cache size across all weights. If a newly activated channel has a higher count than the least-used channel in the cache, we evict the least-used channel. Fig. \ref{F.dynamic_cahce} illustrates an example of our dynamic cache mechanism. For a given sequence, we begin by initializing the usage count of all channels to zero. For the first token, channel index 0 is present in the cache, while channel indices 1, 4, and 6 need to be loaded from flash storage, resulting in a hit ratio of 25\%. For the second token, channel indices 0, 4, and 6 hit in the cache, while only channel index 7 needs to be fetched from flash. Since channel index 1 has the lowest frequency, we replace it with channel index 7, improving the hit ratio to 75\%.

\section{Self-Distillation for Top-K Sparse LLM}

Even with superior quality compared to other sparsity techniques, Top-K activation sparsity still introduces an approximation in active weight selection, especially in high sparsity. Traditional methods such as supervised fine-tuning often fail to recover the performance of the model under high sparsity, as they cannot effectively capture nuanced weight distributions and activation patterns caused by sparsity, leading to a degradation of precision. To address this, we propose \textbf{Top-K sparsity-aware self-distillation}, an extension of quantization and fine-tuning pipelines. It preserves the efficiency benefits of sparsity while substantially reducing computational overhead and enhancing both accuracy and generalization. In practice, it improves performance with only a few to tens of GPU hours on a few thousand samples, and generalizes effectively across different sparsity levels.

\begin{figure}[tb]
    \centering
    \captionsetup[subfigure]{skip=0.3em} 
    \begin{subfigure}[t]{0.45\textwidth}
        \centering
        \includegraphics[width=1\linewidth]{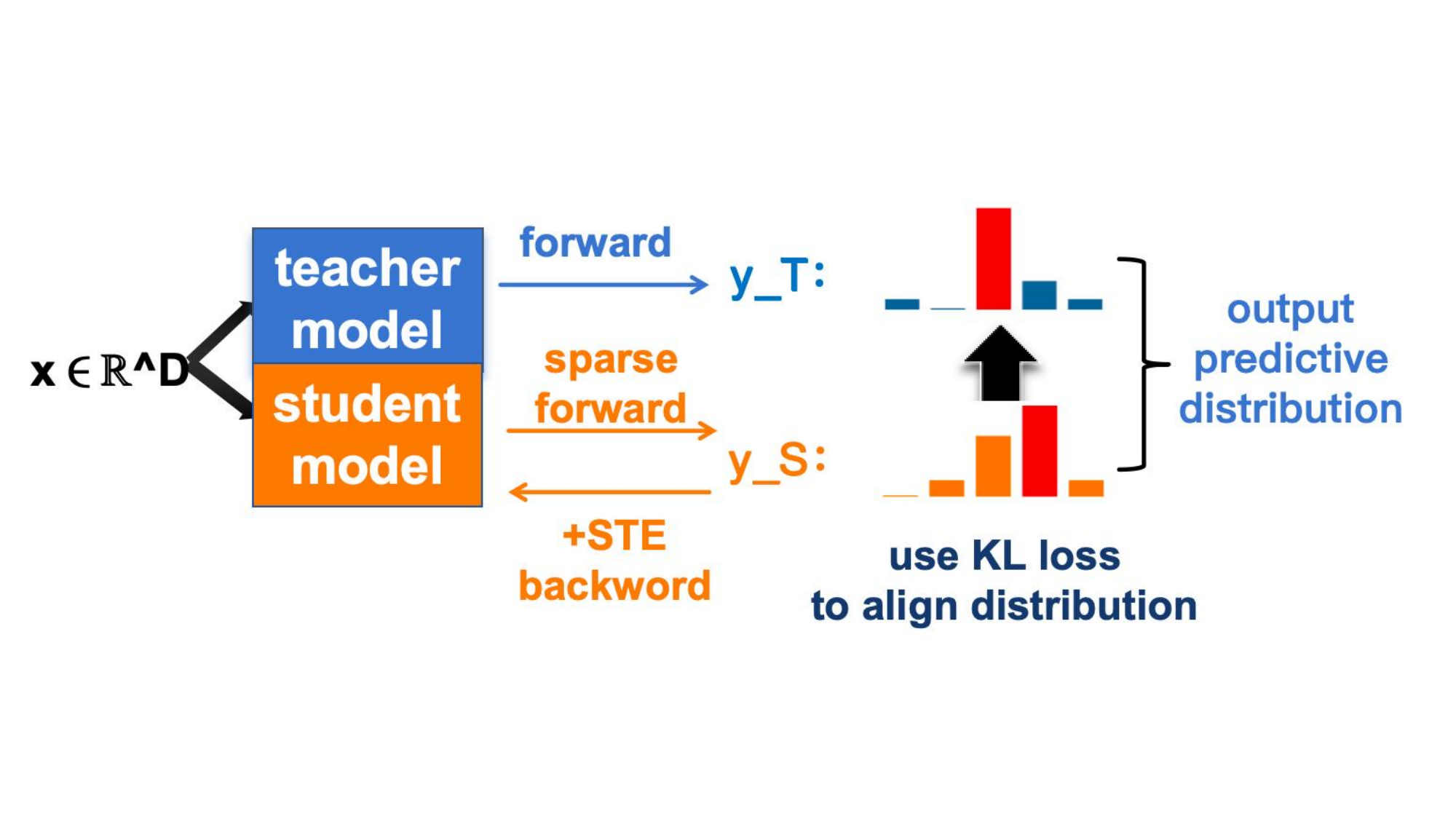}
        \label{F.distill_distribution}
    \end{subfigure}
    \hfill
    \vspace{-3em}
    \begin{subfigure}[t]{0.45\textwidth}
        \centering
        \includegraphics[width=0.45\linewidth]{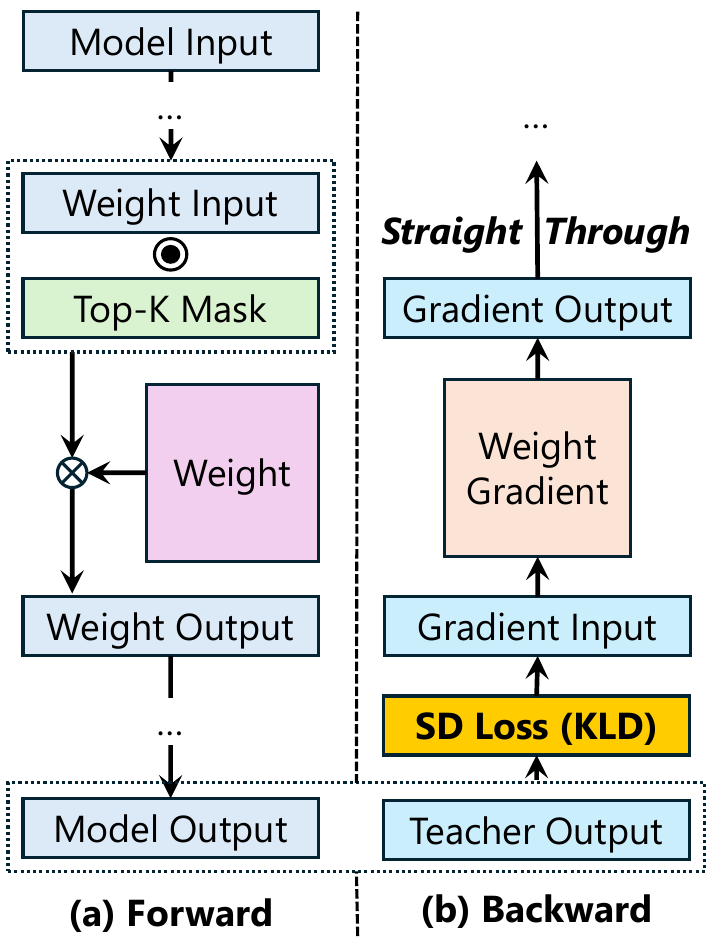}
        \label{F.sd_forward_backward}
    \end{subfigure}
    \caption{The forward and backward of self-distillation between teacher and student.}
    \label{F.sd_combined}
\end{figure}

\textbf{Self distillation.}
As shown in Fig.~\ref{F.sd_combined}, we maintain the outputs of the dense (teacher) and sparse (student) models, using the soft output distribution of the dense model as supervision. This allows the sparse student to capture richer correlations than hard labels and preserve fine-grained distributional details, which is crucial to compensate for information loss induced by Top-K activation sparsity. 

\textbf{KL loss.}
We adopt the \textbf{Kullback-Leibler Divergence (KLD) loss} to measure the discrepancy between the student and teacher distributions:
\begin{align}
    \mathcal{D}_{\text{KL}}(P_T \parallel P_S) &= \sum_{i} P_T(i) \log \frac{P_T(i)}{P_S(i)} \label{eq:kld_loss}
\end{align}
Minimizing $\mathcal{D}_{\text{KL}}$ encourages the sparse model to closely mimic the dense teacher’s distribution, preserving essential weight correlations and improving performance under high sparsity. Our framework is also orthogonal to quantization, as the KL-based distillation loss depends only on the output distribution and thus remains fully compatible with QAT, making it complementary to quantization errors and enabling additional efficiency gains with minimal accuracy loss.

\textbf{Gradient STE.}
Gradient vanishing is a common issue when fine-tuning activation-sparse models, as the sparsity mask sets many elements to zero, preventing gradients from being properly propagated and slowing or even blocking convergence. To mitigate this, we employ \textbf{gradient Straight-Through Estimation (STE)}, which replaces the gradient of the masking operation with an identity function during backpropagation:
\begin{align}
    \text{forward:} & \quad y = \operatorname{Mask}(x) \\
    \text{backward:} & \quad \frac{\partial y}{\partial x} = I
\end{align}
This allows gradients to flow as if the mask contains non-zero values, ensuring sufficient update signals even under high sparsity. Consequently, STE accelerates convergence, enhances training robustness, and helps the model preserve critical weight distributions and activation patterns.

\textbf{Inherent adaptability.}
A key advantage of our self-distillation framework is its inherent \emph{adaptability across sparsity levels}. 

In conventional fine-tuning or distillation pipelines, models at different sparsity ratios typically require separate training processes, which is both time-consuming and computationally expensive. 

In contrast, our method requires only a \emph{single distillation} at a fix sparsity level, where the student is forced to maintain performance under extreme information constraints. The distilled model not only captures the dense teacher’s distributional characteristics but also develops robustness in reconstructing critical features, enabling direct inference across other sparsity levels without additional fine-tuning. This robustness naturally extends with increasing activation budget: model leverages the richer signals without retraining, achieving near-lossless performance over a wide sparsity range. We refer to this property as \textbf{one-distill-all-scale}; as Table~\ref{tab:quant4bit_ppl_transposed} shows, PPL error remains within 1\% even when training and inference sparsity differ by up to $\pm15\%$.

By eliminating the need for repeated training at each sparsity level, this approach greatly reduces overall cost while ensuring consistent accuracy and efficiency in sparse LLM deployment.


%% file: implementation.tex
\section{Implementation}

\sysname is built on \textit{llama.cpp}, a widely-used LLM inference framework for mobile devices. The whole model is stored in flash and only active weight, cached weight, and preloaded weight are in DRAM. This paper is based on the CPU backend of llama.cpp. The big cores execute computations and the little cores execute data loading concurrently. Since decoding speed is memory bandwidth bound, and mobile devices use a unified DRAM among all processors, we believe implementing \sysname on different processors should have similar results. Past work~\cite{weiTMACCPURenaissance2024,wang2025bitnetcppefficientedgeinference} have also demonstrated the superior performance of CPU over NPU for decoding on devices. We thus choose CPU in this paper for implementing convenience.     

\textbf{Flash loading.} To implement cross-layer-group LLM weight loading, we modify the way weight tensors are stored in the GGUF format. Specifically, we save each operator's weights as fundamental tensors organized in a cross-layer-group manner. We utilize \textit{IO uring}, a low-overhead asynchronous I/O mechanism, to read the weights efficiently. In particular, we use the \textit{io\_uring\_prep\_read} and \textit{io\_uring\_submit} functions to asynchronously request reads for active weights. After submitting all read requests, we synchronize the I/O operations using the \textit{io\_uring\_wait\_cqe} function. When reading active weights, we sparsely load different channels into a dense buffer, which helps optimize memory buffer layout for better compactness. Additionally, to ensure compatibility with quantization, we apply a transpose operation to the weights. This allows for complete retrieval of the necessary scaling factors when reading channels, thereby facilitating the quantization.

\textbf{Swapping pipeline.} To implement the active weight swapping pipeline, we first create a dedicated weight loading thread using the \textit{ggml\_thread\_create} function. This thread is bound to a little core of the CPU via the \textit{sched\_setaffinity} function to optimize resource utilization. Synchronization between the weight loading thread and the main computing thread is achieved through atomic semaphores. We use \textit{atomic\_load\_explicit} and \textit{atomic\_store\_explicit} to manage a request signal and a complete signal that facilitates communication between the two threads. The signals operate at the cross-layer-group granularity, ensuring proper execution order between computing and weight loading operations.

\textbf{Caching.} Additionally, we implement the dynamic LLM weight caching, where caching is managed separately for each weight tensor. We use a hash table-based approach to efficiently query cached weight channels and dynamically track their activation frequency during decoding. When loading a new channel, we replace the least frequently activated channel, updating its index pointer in the hash table accordingly. Furthermore, we develop a kernel for generating active channel indices. This kernel maintains activation thresholds corresponding to different LLM sparsity levels. Before each activation step, it determines whether a channel should be activated based on the appropriate threshold. 

\textbf{Self-distillation.} In order to implement the sparsity-aware self-distillation, we develop a plug-and-play sparse module in BitDistiller \cite{du2024bitdistiller}, an open-source framework for quantization-aware LLM distillation. Specially, we insert an activation sparsity module before each LLM weight computation. This module preloads a sparsity threshold for the activations and generates a Top-K mask at inference time by comparing the activations against the threshold. During backpropagation, we incorporate a gradient STE layer for each LLM weight. In addition, we implement the KLD loss function. In our self-distillation experiments, we use a sub-dataset from C4 dataset, with each epoch containing approximately 50K data samples (10B tokens). Full self-distillation comprises two epochs, with a learning rate of $1\times10^{-6}$ and 4-bit quantization. On 4$\times$80G-A100, it takes approximately 10 hours for an LLM.

Overall, \sysname comprises 3762 new lines of C++ code and thousands lines of Python code.

%% file: evaluation.tex
\section{Evaluation}
\label{evaluation}

We evaluate \sysname on both end-to-end and technique performance, compared to several baselines. The evaluation setup is as follows:

\subsection{Evaluation setup}

\textbf{Hardware devices.} As shown in Table \ref{T.hardware_device}, we evaluate \sysname on three mobile devices, covering a range from high-end to low-end. For clarity, we label the three devices as Device 1, Device 2, and Device 3. 

\textbf{Models.} To assess end-to-end performance, we test popular LLMs, including the Llama and Mixtral series, with model sizes ranging from 7B to 56B parameters. All LLMs undergo 4-bit quantization using \textit{Q4\_0}, a widely used technique that has minimal impact on accuracy. For the technique evaluation, we extract and use eight layers from the original LLM.

\textbf{Baselines.} We compare \sysname against \textit{llama.cpp} in terms of decoding speed and memory usage. For perplexity and accuracy evaluation, we use the original LLM, ProSparse, and TEAL as baselines. ProSparse and TEAL represent state-of-the-art ReLU-sparse and Top-K-sparse LLMs, respectively.

\textbf{Measurement.} Our evaluation focuses on decoding speed, perplexity, accuracy, latency, hit rate, memory cost, power, and energy consumption. We use the \textit{clock\_gettime} function to record start and end timestamps, computing latency as the difference between them. We measure the total number of decoded tokens and the total decoding time, calculating speed as $N_{tokens}/Latency$. We use \textit{lm-eval-harness}, a widely used LLM evaluation framework, to measure perplexity on the WikiText-2 dataset and accuracy on five downstream tasks: 5-shot MMLU, 5-shot GSM8K, 25-shot ARC Challenge, 25-shot ARC Easy, and 0-shot PIQA. We track cache hits and misses, computing the hit rate as $N_{hit}/(N_{hit}+N_{miss})$. We analyze memory cost using the Android Studio Profiler. We obtaine current and voltage values by reading system files (voltage\_now and current\_now) to calculate power consumption. These values are collected every 0.5 seconds on average, and we use the decoding latency to compute the overall energy consumption.

\begin{table}[t]
  \footnotesize
  \caption{The hardware devices for evaluation.}
  \label{T.hardware_device}
  \begin{tabular}{cccc}
    \toprule
    Device & CPU & Memory & \makecell[c]{Flash (MaxBW)}\\
    \midrule
    OnePlus 12 & X4+A720+A520 & 16GB & \makecell[c]{UFS 4.0 (5.8 GB/s)} \\
    Pixel 6 & X1+A76+A55 & 8GB & \makecell[c]{UFS 3.1 (4.2 GB/s)}\\
    Infinix ZERO 30 & A76+A55 & 8GB & \makecell[c]{UFS 2.2 (3.6 GB/s)}\\
  \bottomrule
\end{tabular}
\end{table}

\begin{figure}[t]
    \centering
    \begin{subfigure}[t]{0.24\textwidth}  
        \centering 
        \includegraphics[width=\linewidth]{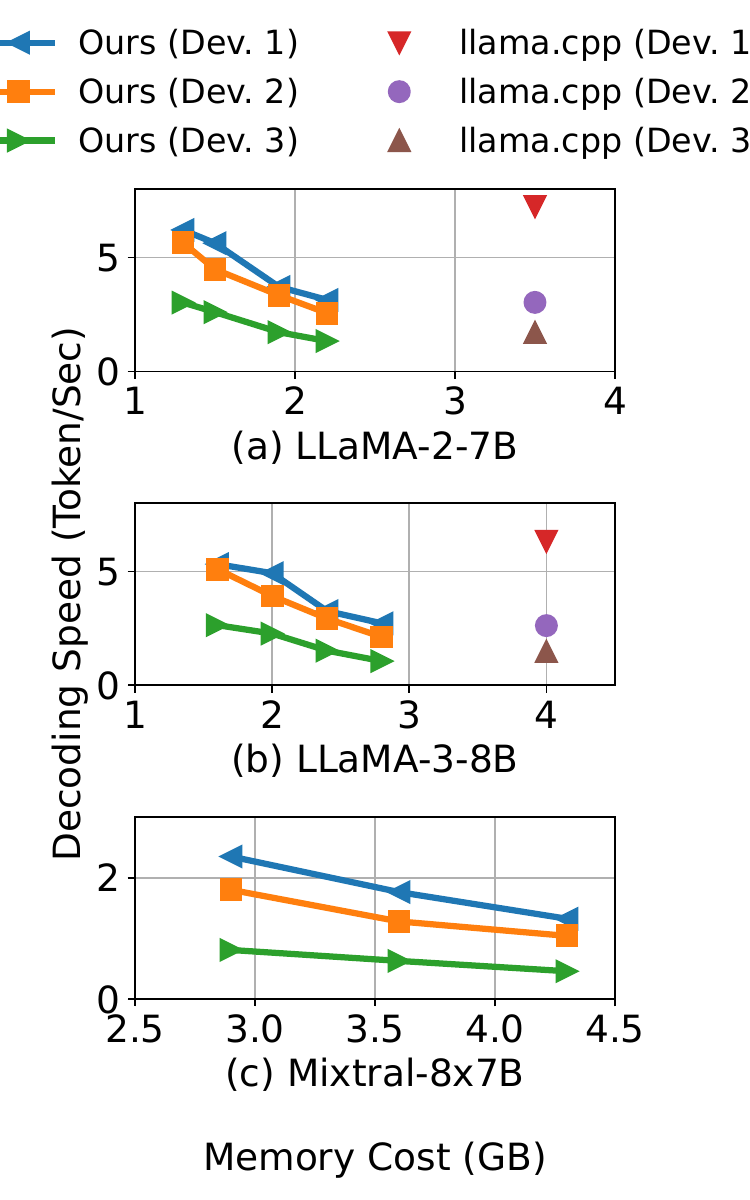}
        \caption{}
        \label{F.end_to_end_decoding_speed_and_memory_cost}
    \end{subfigure}
    \begin{subfigure}[t]{0.20\textwidth}
        \centering
        \includegraphics[width=\linewidth]{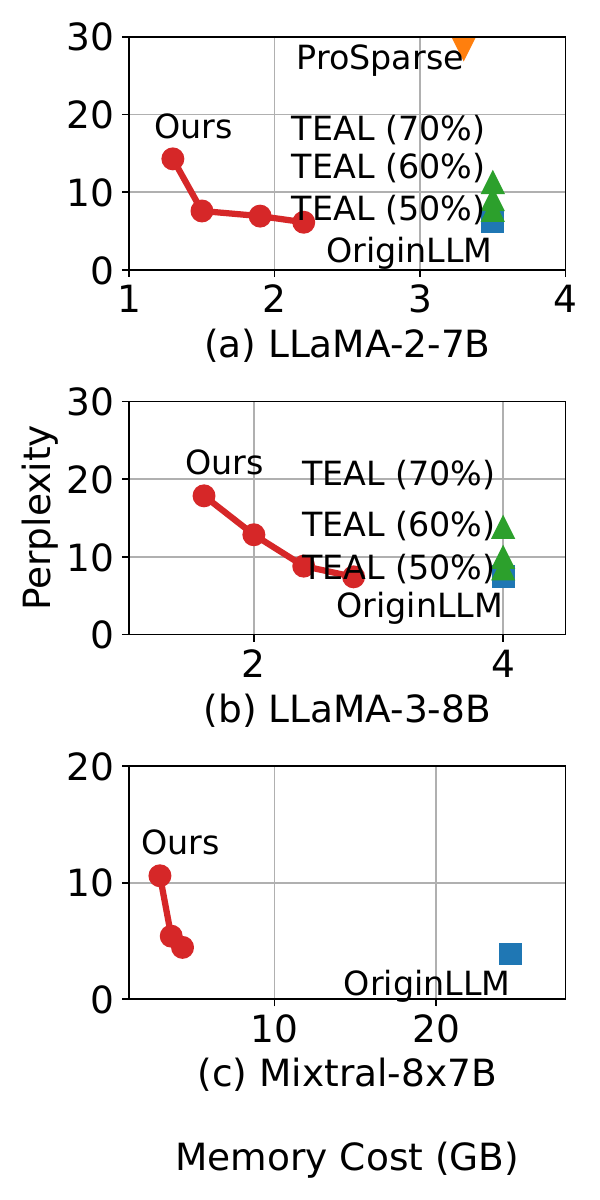}
        \caption{}
        \label{F.end_to_end_ppl_and_memory_cost}
    \end{subfigure}
    \vspace{0em}
    \caption{The end-to-end decoding speed, perplexity and memory cost of three LLMs compared with baselines on various devices. Each point represents a sparsity ratio: from left to right 0.8, 0.7, 0.6, 0.5. Since decoding is memory bound, latency increases with less sparsity and more memory cost.}
    \vspace{-0.5em}
\end{figure}

\subsection{End-to-end performance}

\textbf{Decoding speed.} We first evaluate the decoding speed of different LLMs across various devices under different memory cost conditions as illustrated in Fig. \ref{F.end_to_end_decoding_speed_and_memory_cost}. For Device 2 and Device 3, using the LLaMA-2-7B model, we achieved the same performance as the full-weight memory setting while reducing memory cost by 40\%. When reducing memory cost by 75\%, our method achieved a 1.9× and 1.5× speedup compared to the full-weight in-memory setting on Device 2 and Device 3, respectively. The speedup is primarily due to our computing-loading pipeline, which enables higher decoding speed even under lower memory cost constraints. However, on Device 1, when using 60\% of the memory cost, our performance dropped by 54\% compared to the full-weight memory setting. This is because the CPU compute bandwidth of Device 1 is significantly higher than its flash read bandwidth, making the pipeline constrained by flash bandwidth. Nonetheless, at 75\% memory cost, our method was able to achieve a decoding speed of 5.9 tokens per second.

For the Mixtral model, we successfully enable decoding under 6GB of memory. When the memory cost was 4.3GB, the decoding speed on Device 1, Device 2, and Device 3 was 1.3, 1.0, and 0.4 tokens per second, respectively. As the memory cost was reduced to 2.9GB, the performance improved to 2.3, 1.8, and 0.8 tokens per second, achieving a 1.8× to 2.0× speedup across the three devices.

\textbf{Perplexity and Downstream tasks.} 
Our method demonstrates that large language models can maintain low perplexity under significantly reduced memory costs, e.g., achieving performance comparable to the full-weight setting for LLaMA-2-7B and LLaMA-3-8B at only 60\% memory usage in Fig. \ref{F.end_to_end_ppl_and_memory_cost}, and matching the Mixtral-8x7B baseline (24.6GB) with just 4.4GB. While perplexity increases under more aggressive sparsity, our self-distillation strategy effectively alleviates performance degradation, enabling consistent improvements over TEAL across five downstream tasks, with gains up to 10.98\% at 70\% sparsity and average improvements ranging from 2.64\% to 10.21\% across sparsity levels in Table~\ref{T.performance_llama_3_8b}.

\textbf{ablation study for self-distillation.}
To validate the effectiveness of each component in our framework, we conducted ablation studies focusing on the gradient straight-through estimator (STE) and self-distillation techniques. We carried out experiments on the Llama-3-8B model, comparing performance under different configurations, including: 1. removing STE; 2. replacing self-distillation with full fine-tuning. As shown in Table~\ref{T.performance_comparison}, our framework’s components improve model performance across different sparsity levels. 

\begin{figure}
    \centering
    %
    \includegraphics[width=1\linewidth]{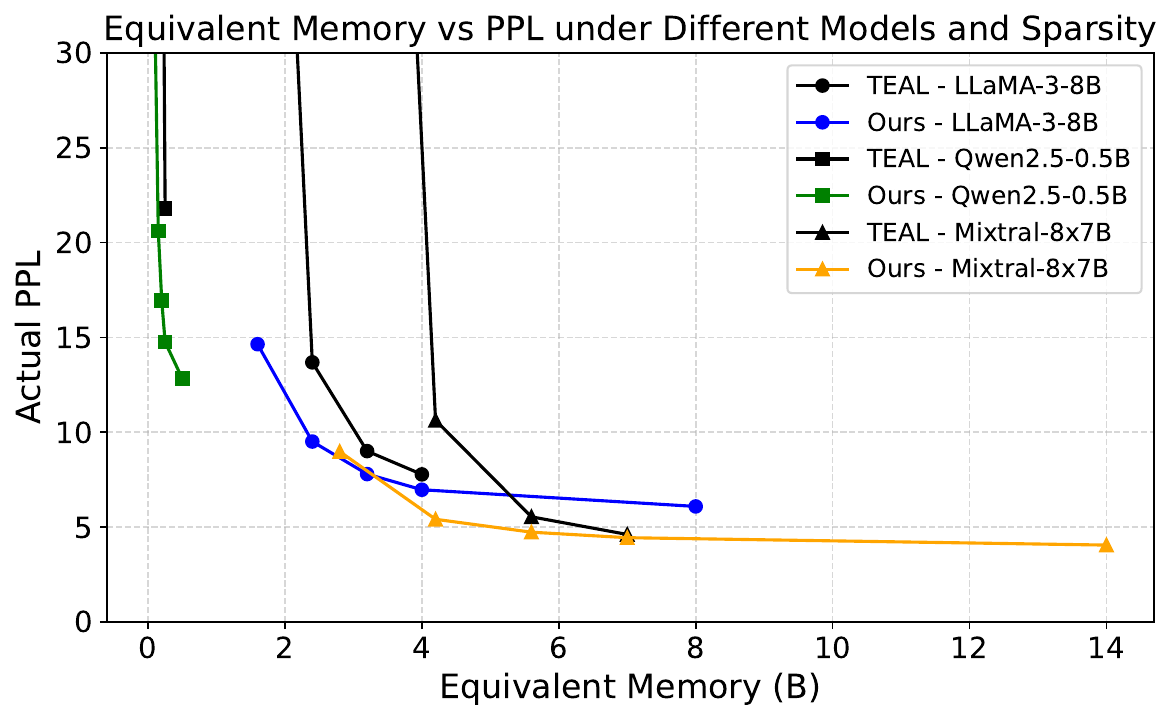}
    \vspace{-1em}
    \caption{Pareto frontier of actual runtime memory vs. PPL for TEAL and our self-distillation.}
    \label{F.model_pareto_front}
\end{figure}

\textbf{different models for self-distillation.}
We evaluated the proposed self-distillation framework across diverse model architectures and visualized the results in equivalent memory–performance plots. The method demonstrates strong generality: from standard 7B models to highly compressed Qwen2.5-0.5B and sparse MoE architectures, it consistently delivers significant sparsity-driven performance gains. Unlike conventional compression, our approach ensures predictable accuracy loss while achieving strict acceleration. As illustrated in Figure \ref{F.model_pareto_front}, our results consistently lie on the Pareto frontier of equivalent memory and performance, underscoring the framework’s adaptability for practical deployment.

\begin{table}[t]
\centering
\caption{End-to-End PPL results under Varying Sparsity Levels of Meta-LLaMA-3-8B (distillation under Fixed-Sparsity), using 4-bit quantization.}
\label{tab:quant4bit_ppl_transposed}
\resizebox{\linewidth}{!}{%
\begin{tabular}{lccccc}
\toprule
\textbf{Method} & \textbf{0\%} & \textbf{50\%} & \textbf{60\%} & \textbf{70\%} & \textbf{80\%} \\
\midrule
TOP-K              & 6.6836    & 8.1950    & 10.0121   & 15.9046   & 96.3015   \\
Ours               & —         & 7.4510    & 8.3216    & 10.2442   & 16.1081   \\
Ours distill on 50\% & —       & 7.4510    & 8.5625    & 11.8981   & 41.5303   \\
Ours distill on 60\% & —       & 7.4636    & 8.3216    & 10.8789   & 27.2720   \\
Ours distill on 70\% & —       & 7.8163    & 8.4440    & 10.2442   & 19.6981   \\
Ours distill on 80\% & —       & 9.3462    & 9.8854    & 11.1767   & 16.1081   \\
\bottomrule
\end{tabular}%
}
\end{table}

\begin{table}
\footnotesize
\centering
\caption{PPL and Downstream Task Accuracy of LLaMA-3-8B.We use TEAL as our baseline TOP-K method.}
\label{T.performance_llama_3_8b}
\resizebox{\linewidth}{!}{%
\begin{tabular}{lcccccc}
\toprule
\textbf{Method} & \textbf{PPL } & \textbf{MMLU} & \textbf{GSM8K} & \textbf{ARC-C} & \textbf{ARC-Easy} & \textbf{PIQA} \\
\midrule
Origin (0\%) & 6.0874 & 65.16\% & 50.87\% & 54.95\% & 83.96\% & 80.74\% \\
\midrule
TOP-K (50\%) & 7.7762 & 59.21\% & 32.30\% & 49.32\% & 81.40\% & 78.45\% \\
TOP-K (60\%) & 9.0042 & 51.67\% & 17.76\% & 45.56\% & 76.98\% & 76.01\% \\
TOP-K (70\%) & 13.6816 & 36.53\% & 3.34\% & 33.70\% & 66.60\% & 70.30\% \\
TOP-K (80\%) & 73.1400 & 25.69\% & 1.67\% & 21.08\% & 38.89\% & 56.80\% \\
\midrule
Ours (50\%) & \textbf{6.9677} & \textbf{61.41\%} & \textbf{38.89\%} & \textbf{52.13\%} & \textbf{81.48\%} & \textbf{79.98\%} \\
Ours (60\%) & \textbf{7.7935} & \textbf{57.01\%} & \textbf{28.58\%} & \textbf{49.57\%} & \textbf{79.46\%} & \textbf{77.80\%} \\
Ours (70\%) & \textbf{9.5079} & \textbf{47.51\%} & \textbf{12.89\%} & \textbf{41.64\%} & \textbf{74.12\%} & \textbf{77.31\%} \\
Ours (80\%) & \textbf{14.6401} & \textbf{29.40\%} & \textbf{2.05\%} & \textbf{32.34\%} & \textbf{62.30\%} & \textbf{69.10\%} \\
\bottomrule
\end{tabular}%
}
\end{table}

\begin{figure}[!t]
    \centering
    \includegraphics[width=0.50\textwidth]{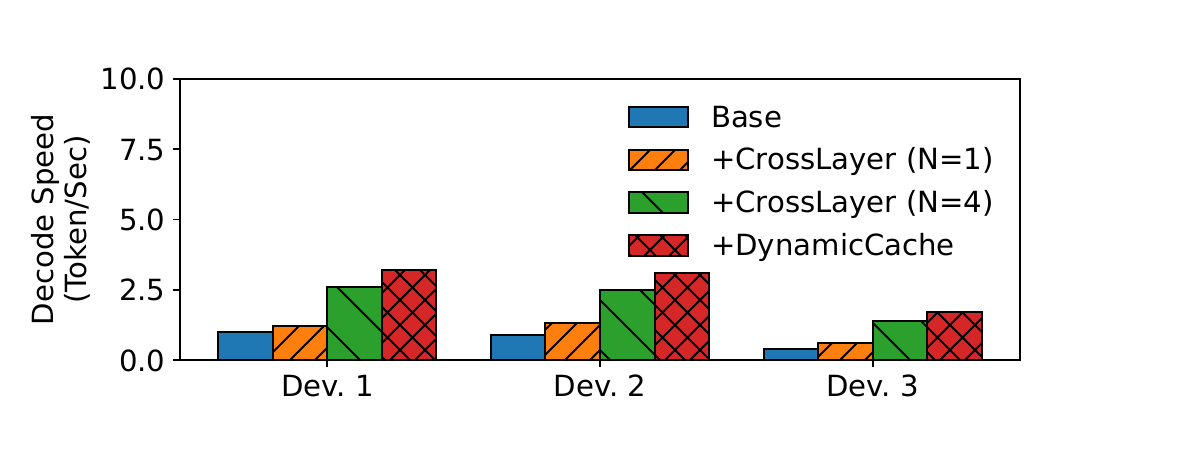}
    \vspace{-1em}
    \caption{The decode speed improvement of LLaMA-2-7B model on three devices by each technique.}
    \label{F.technique_breakdown_on_decode_speed}
\end{figure}

\subsection{Technique breakdown}


To validate the effectiveness of our system’s techniques, we conduct ablation studies and standalone tests for each component, evaluating their impact on decoding speed, perplexity, and hit rate.

\textbf{Cross-layer-group pipeline.} First, we examine the effect of the cross-layer-group pipeline on decoding speed, as shown in Fig. \ref{F.technique_breakdown_on_decode_speed}. We used a 60\% sparsity LLaMA-2-7B model and tested it across three devices. Our baseline consisted of serial computation and memory reads. Experimental results show that when the layer number in a cross-layer group is set to 1, the average speedup across all three devices is 10\%. However, increasing the layer number to 4 results in a 120\% performance improvement, as it enhances the efficiency of flash memory reads. Finally, with the addition of Dynamic Cache, our method achieves 2$\times$, 2.3$\times$, and 3$\times$ speedups over the baseline on the three devices, respectively.

\begin{figure}[!t]
    \centering
    \begin{subfigure}[t]{0.23\textwidth}
        \centering
        \includegraphics[width=\textwidth]{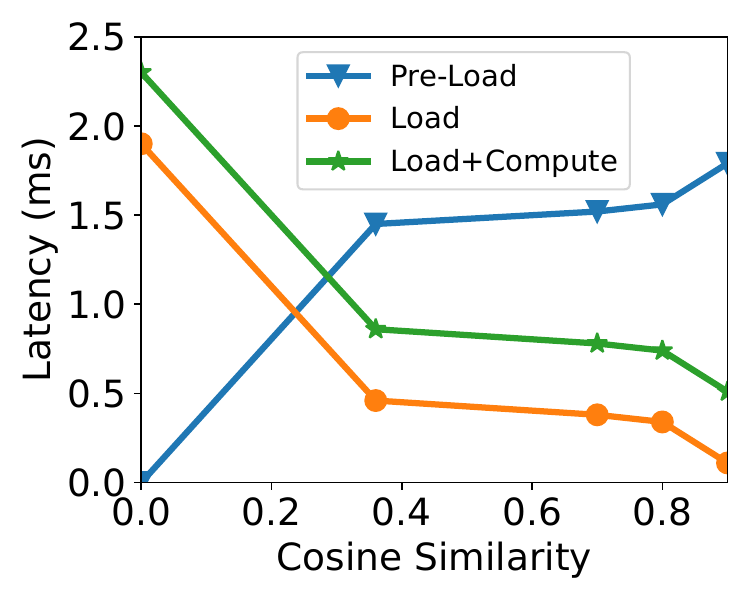}
        \caption{}
        \label{F.cosine_similarity_and_loading_time}
    \end{subfigure}
    \begin{subfigure}[t]{0.24\textwidth}
        \centering
        \includegraphics[width=\textwidth]{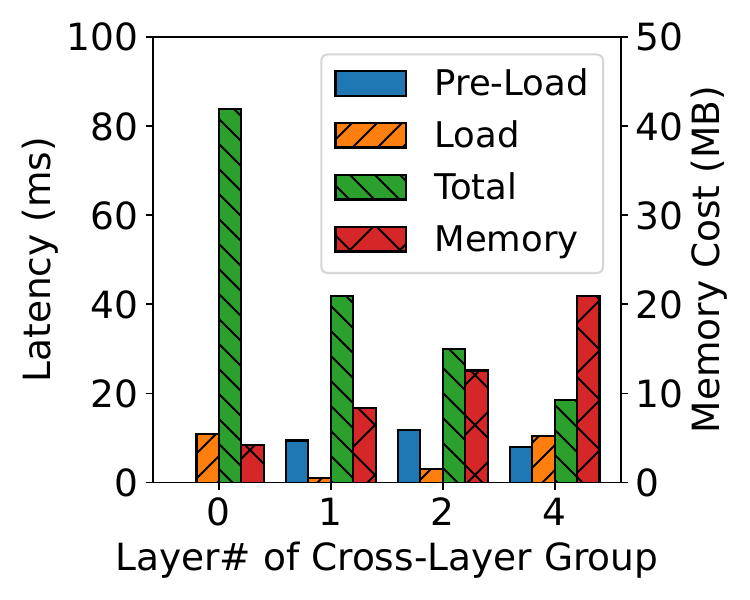}
        \caption{}
        \label{F.cross_layer_length_and_latency_and_memory}
    \end{subfigure}
    \vspace{0em}
    \caption{The performance and memory cost of cross-layer loading.}
    \label{F.cross_layer_loading_performance}
\end{figure}

To further understand the benefits and overhead of each technique, we conducted individual experiments for detailed analysis. As shown in Fig. \ref{F.cross_layer_loading_performance}, we evaluated the trade-offs of cross-layer loading. In Fig. \ref{F.cross_layer_loading_performance}(a), we measured the loading and preloading overhead for a single layer when the layer number in a cross-layer group is set to 1, under different cosine similarity values. The results show that when cosine similarity is lower than 0.2, the preload latency is lower than the on-demand load latency. However, when cosine similarity exceeds 0.4, the on-demand load latency becomes lower than the preload latency. Since the cosine similarity of most layers is above 0.8, our cross-layer approach effectively overlaps preloading and computation, optimizing performance.



\begin{figure}[!t]
    \centering
    \begin{subfigure}[t]{0.23\textwidth}
        \centering
        \includegraphics[width=\textwidth]{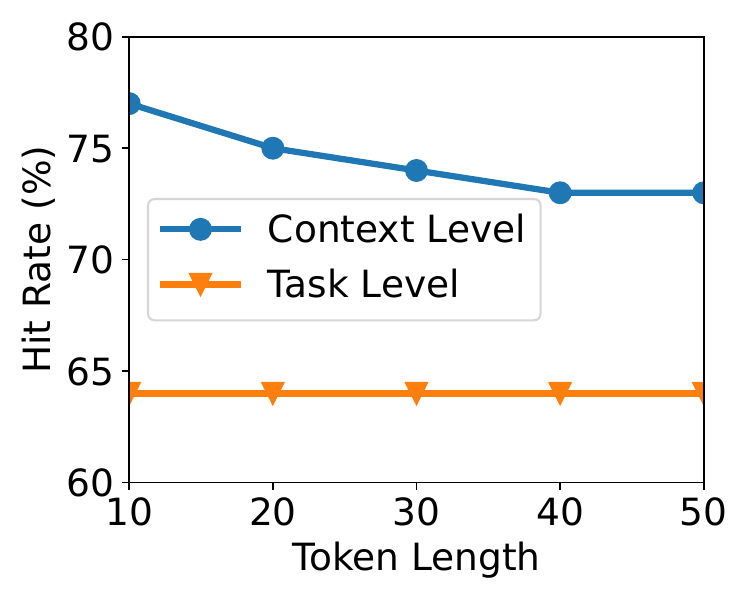}
        \caption{}
        \label{F.sample_count_and_cache_hit_rate.pdf}
    \end{subfigure}
    \begin{subfigure}[t]{0.23\textwidth}
        \centering
        \includegraphics[width=\textwidth]{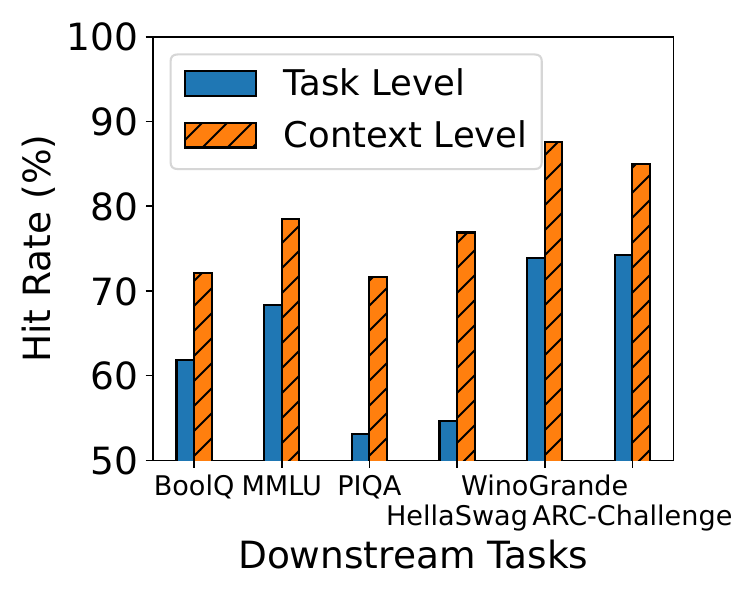}
        \caption{}
        \label{F.downstream_tasks_and_cache_hit_rate}
    \end{subfigure}
    \vspace{0em}
    \caption{The performance of task-level and context-level cache.}
    \label{F.cache_performance}
\end{figure}

In Fig. \ref{F.cross_layer_loading_performance}(b), we evaluate an 8-layer decoder of LLaMA-2-7B, measuring preload, load, and total latency as well as memory cost under different layer numbers in a cross-layer group. When the layer number is 0, computation and flash loading occur sequentially, leading to high total latency. When the layer number increases to 1, computation begins to overlap with preloading, reducing total latency by 52\%. As the layer number further increases to 4, improved preload efficiency enables a 4.1× speedup compared to the size 0 setting. However, increasing the layer number also leads to higher memory cost, introducing additional overhead. Overall, increasing the layer number in a cross-layer group effectively enhances decoding performance, while the additional memory overhead remains relatively low.

\textbf{Contextual caching policy.} 
Fig.~\ref{F.cache_performance} compares context-level and task-level caches. On BoolQ, when token length=10, the context-level cache achieves a 77\% hit rate, 13\% higher than task-level. As length increases to 40, the hit rate slightly drops to 74\% but still remains 10\% higher. Across downstream tasks (Fig.~\ref{F.cache_performance}b), task-level hit rate varies between 54–74\%, while context-level consistently adapts, yielding an average 12\% improvement.  

\begin{figure}[!t]
    \centering
    \includegraphics[width=0.50\textwidth]{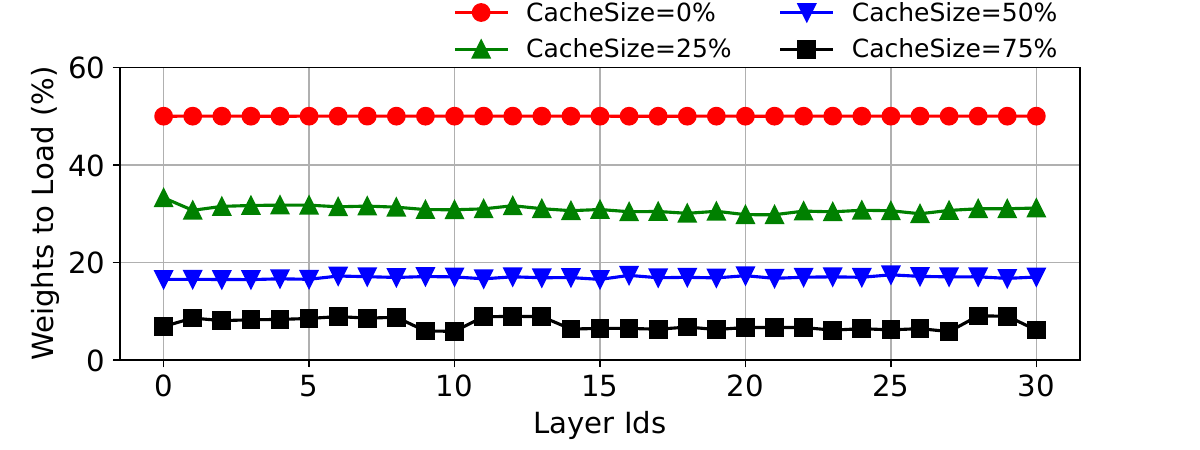}
    \vspace{-1em}
    \caption{The rate of attention Q/K/V weights to load of LLaMA-2-7B model with 50\% sparsity under various cache sizes.}
    \label{F.cache_rate_and_load_rate}
\end{figure}

\textbf{Cache efficiency.} 
As shown in Fig.~\ref{F.cache_rate_and_load_rate}, enlarging cache size significantly reduces flash access. With 50\% cache, flash operations shrink to 18\% of weights, giving a 5.2× reduction in memory access compared to full loading. Larger cache further improves hit rate but also increases memory footprint; therefore, we adjust cache size dynamically based on available device memory.

\begin{figure}[tb]
    \centering
    \begin{subfigure}[t]{0.23\textwidth}
        \centering
        \includegraphics[width=0.8\textwidth]{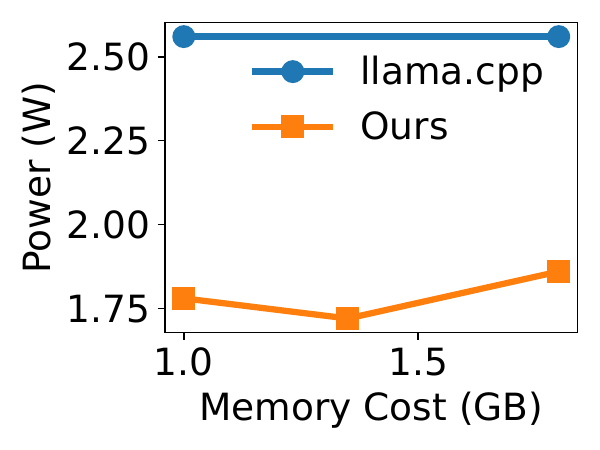}
        \caption{}
        \label{F.power}
    \end{subfigure}
    \begin{subfigure}[t]{0.23\textwidth}
        \centering
        \includegraphics[width=0.8\textwidth]{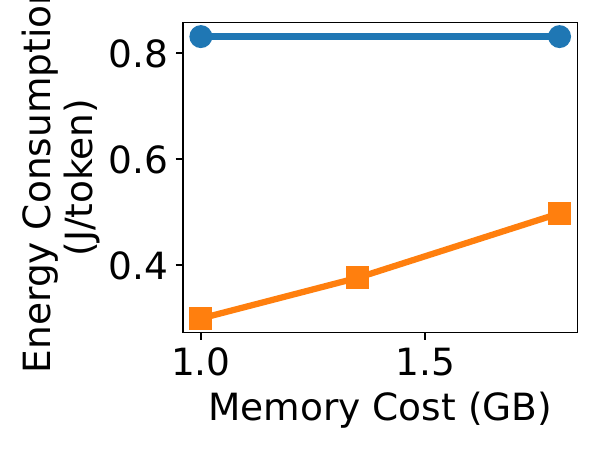}
        \caption{}
        \label{F.energy_consumption}
    \end{subfigure}
    \vspace{-1em}
    \caption{The power and energy consumption of \sysname and baseline.}
    \label{F.power_and_energy_consumption}
\end{figure}

\begin{table}
\footnotesize
\centering
\caption{Llama-3-8B ablation studies}
\label{T.performance_comparison}
\resizebox{\linewidth}{!}{%
\begin{tabular}{lcccccc}
\toprule
\textbf{Method} & \textbf{PPL} & \textbf{MMLU} & \textbf{GSM8K} & \textbf{ARC-C} & \textbf{ARC-Easy} & \textbf{PIQA} \\
\midrule
\textbf{Ours (50\%)} & \textbf{6.9677} & \textbf{61.41\%} & \textbf{38.89\%} & \textbf{52.13\%} & \textbf{81.48\%} & \textbf{79.98\%} \\
Ours-Distill (50\%) & 7.4872 & 59.67\% & 32.83\% & 50.94\% & 81.82\% & 78.89\% \\
Ours-STE (50\%) & 7.0660 & 60.78\% & 37.45\% & 49.91\% & 81.65\% & 79.54\% \\
\midrule
\textbf{Ours (60\%)} & \textbf{7.7935} & \textbf{57.01\%} & \textbf{28.58\%} & \textbf{49.57\%} & \textbf{79.46\%} & \textbf{77.80\%} \\
Ours-Distill (60\%) & 8.2635 & 55.73\% & 24.64\% & 50.68\% & 80.39\% & 77.53\% \\
Ours-STE (60\%) & 8.1517 & 55.27\% & 23.58\% & 47.78\% & 78.20\% & 77.91\% \\
\midrule
\textbf{Ours (70\%)} & \textbf{9.5079} & \textbf{47.51\%} & \textbf{12.89\%} & \textbf{41.64\%} & \textbf{74.12\%} & \textbf{77.31\%} \\
Ours-Distill (70\%) & 9.9149 & 45.29\% & 10.77\% & 42.41\% & 73.99\% & 75.41\% \\
Ours-STE (70\%) & 11.1969 & 37.72\% & 3.87\% & 35.92\% & 68.69\% & 73.18\% \\
\midrule
\textbf{Ours (80\%)} & \textbf{14.6401} & \textbf{29.40\%} & \textbf{2.05\%} & \textbf{32.34\%} & \textbf{62.30\%} & \textbf{69.10\%} \\
Ours-Distill (80\%) & 37.7404 & 24.44\% & 1.90\% & 20.14\% & 36.95\% & 57.67\% \\
Ours-STE (80\%) & 23.2097 & 25.96\% & 1.59\% & 22.87\% & 52.74\% & 64.85\% \\
\bottomrule
\end{tabular}%
}
\end{table}

\subsection{Power and energy consumption}

We evaluate power and energy efficiency of \sysname on Device 1 (Fig. \ref{F.power_and_energy_consumption}). \sysname reduces average power consumption by 27.34\% compared to llama.cpp due to reduced computation wait time in the overlap pipeline, and further lowers energy per token as memory cost decreases, achieving up to 53\% reduction at 1.3GB memory usage.

\vspace{0.3em}

%% file: related_works.tex
\section{Related Works}

\textbf{Sparsity in LLMs.} Sparsity in LLMs has been the focus of many research efforts. Mirzadeh et al.~\cite{mirzadeh2023relustrikesbackexploiting} propose replacing the ReLU activation function in LLMs to reduce computation and weight transfer. HiRE~\cite{l2024hirehighrecallapproximate} introduces high-recall approximate Top-K estimation. Prosparse~\cite{prosparse} leverages the sparsity of ReLU and gated branches in FFNs to predict model sparsity. Q-Sparse~\cite{wang2024qsparselargelanguagemodels} trains sparse LLMs from scratch, while TEAL~\cite{liu2025trainingfreeactivationsparsitylarge} applies magnitude-based sparsity without retraining. However, these methods either depend on ReLU-based architectures or lack mechanisms to recover accuracy under high sparsity. InfiniGen\cite{lee2024infinigenefficientgenerativeinference}, FlexGen~\cite{sheng2023flexgenhighthroughputgenerativeinference} and related work primarily focus on \textbf{KV cache optimization}, KV cache dominants the LLM memory usage for long context scenarios (>32K tokens). while we target \textbf{weight} memory optimization: usually determined by LLM's weights.

\textbf{Efficient LLM inference system.} Several system-level efforts focus on exploiting sparsity for efficient inference. DejaVu~\cite{pmlr-v202-liu23am} predicts contextual sparsity with lightweight algorithms, Alizadeh et al.~\cite{alizadeh2024llm} optimize inference on limited-memory devices, and PowerInfer~\cite{10.1145/3694715.3695964} (and its extension PowerInfer-2) design CPU–GPU hybrid engines. These works primarily target \textit{ReLU-based models} and FFN layers, often relying on heavy predictors (GB-level memory) to skip zero activations. Yet modern LLMs such as LLaMA and Mixtral adopt non-ReLU activations for accuracy~\cite{touvron2023llama}, limiting the applicability of these methods. LLM-in-Flash~\cite{alizadeh2024llmflashefficientlarge} streams weights from flash with fine-grained prefetching to reduce DRAM usage, but its efficiency is limited by flash bandwidth and latency, especially for compute-intensive layers. 

\textbf{Our distinction.}\sysname eliminates the ReLU dependency and predictor overhead by targeting \textit{all weights} (Attention and FFN) in modern non-ReLU LLMs. It targets \textit{all weights} (both Attention and FFN) and eliminates the need for predictors. It introduces (1) cross-layer active weight preloading, generalizing cross-layer similarity, and (2) sparsity-aware self-distillation to recover accuracy under high sparsity. Combined with an LFU-based cache driven by activation statistics, our approach consistently achieves higher hit rates (e.g., $>$70\% vs. $\sim$55\%) and ensures strict memory budgets, enabling reliable edge deployment. 

\textbf{Static pruning techniques.} Static pruning and quantization are established methods for compressing large language models (e.g., CFSP \cite{wang-etal-2025-cfsp}, DB-LLM \cite{chen2024dbllmaccuratedualbinarizationefficient}, and RIA \cite{biswas2025regularizationbasedframeworkquantizationfault}). Although effective at reducing model size and computation, these static approaches require offline processing of model weights, which limits flexibility for dynamic tasks. Our system, \sysname, is not only compatible with these static techniques but also uniquely supports dynamic processing to address this challenge.

\vspace{-0.5em}

%% file: conclusion.tex
\section{Conclusion}
This paper proposes the first LLM inference system on mobile devices that supports adaptive DRAM usage, in order to scale up the deployable model size. It is based on the idea of active weight swapping between DRAM and flash, integrating three novel techniques: cross-layer active weight preloading, sparsity-aware self-distillation, and active weight swapping pipeline. It achieves the inference performance-cost Pareto frontier compared to other efficiency optimization methods. This paper breaks the DRAM limitation for LLM deployment, opening up the new opportunity of server-level LLMs deployment on mobile devices.